\documentclass[runningheads]{llncs}
\usepackage{amsmath}
\usepackage{graphicx}
\usepackage{todonotes}
\usepackage{float}
\usepackage[export]{adjustbox}
\usepackage{siunitx}
\PassOptionsToPackage{hyphens}{url}\usepackage{hyperref}
\usepackage[belowskip=1pt,aboveskip=1pt]{subcaption}
\usepackage[belowskip=-12pt,aboveskip=3pt]{caption}

\usepackage{siunitx}
 
\begin{document}
\title{Contact Planning for the ANYmal Quadruped Robot using an Acyclic Reachability-Based Planner \thanks{This research is supported by the UKRI and EPSRC (EP/R026084/1, EP/R026173/1, EP/S002383/1) and the EU H2020 project MEMMO (780684). This work has been conducted as part of ANYmal Research, a community to advance legged robotics.}}
\titlerunning{Contact Planning for ANYmal}
%
\author{Mathieu Geisert\inst{1} \and 
Thomas Yates\inst{1} \and
Asil Orgen \inst{2}  \and \\
Pierre Fernbach \inst{3} \and
Ioannis Havoutis\inst{1}} 
\authorrunning{M. Geisert et al.}
%
\institute{Oxford Robotics Institute - University of Oxford, Oxford, United Kingdom \and
Faculty of Engineering and Natural Sciences - Sabanci University, Istanbul, Turkey \and
Laboratoire d'Analyse et d'Architecture Syst\`eme - CNRS, Toulouse, France
}
\maketitle              
\begin{abstract}

Despite the great progress in quadrupedal robotics during the last decade, 
selecting good contacts (footholds) in highly uneven and cluttered environments
still remains an open challenge.
This paper builds upon a state-of-the-art approach, already successfully used 
for humanoid robots, and applies it to our robotic platform; 
the quadruped robot ANYmal.
The proposed algorithm decouples the problem into two subproblems: first a guide trajectory for the robot is generated, then contacts are created along this trajectory.
Both subproblems rely on approximations and heuristics that need to be tuned.
The main contribution of this work is to explain how this algorithm has been retuned to work with ANYmal and to show the relevance of the approach with 
a variety of tests in realistic dynamic simulations.

\end{abstract}

\section{Introduction}

Many quadruped robots have shown great control capabilities while moving on difficult terrains such as grass, ice or stairs. However, many examples mostly rely on the intrinsic robustness of quadruped robots and reactive locomotion
approaches based on body velocity estimation, to reject unpredicted
perturbations. Navigating through highly uneven and cluttered environments, often with only a small set of potential footholds, is still an open problem. Some of the the most impressive results on the problem come from the DARPA Learning Locomotion project using the LittleDog robot \cite{Kalakrishnan2010}. This small quadruped was able to navigate through terrains with rocks of a size comparable to its body. However, such performance has still to be reproduced on bigger quadruped robot platforms.

In this paper, we present an approach to automatically compute a contact plan on challenging and uneven terrains. This is only the first step towards our goal to build a generic framework, that can produce consistently good plans for most environments that a quadruped robot can encounter. 
However, this is an important step since several papers have shown that once a feasible footstep plan has been generated, a stable Whole-Body trajectory can be computed in real-time \cite{Carpentier2018,Ponton2016a}.

\subsection{Related Work}

Planning contacts is a difficult problem as the algorithm needs to simultaneously take into account the capabilities of the robot (kinematics and dynamics) and the shape of the terrain (non-smooth and cluttered).
On one hand, the non-continuity and non-convexity resulting from uneven terrain and obstacles make the problem difficult to solve using optimization techniques.
On the other hand, the number of degrees of freedom and the contact constraints make the problem difficult to solve with sampling-based methods.
Moreover, checking for collisions between the robot and the environment make this problem even more difficult to solve fast enough to result in a reactive motion planner.

For simple cases like flat terrains, optimization techniques are able to correctly solve, simultaneously, the motion of the main body of the robot and its footstep placements \cite{Herdt2010a,Naveau2016}. Using more complex models and solvers, the problem can be reformulated to solve motion on other terrains \cite{2017icra_mastalli,Winkler2018}. However, this algorithm is doomed to fall into local minima, e.g. ignoring intermediate steps on stairs or trying to jump over impassable obstacles. Such behaviors can be reduced by first relaxing the complementary constraint of contacts then slowly converging back to the initial problem  \cite{Mastalli2016ICRA,Posa2016,Mordatch2012}. Alternatively, one can decompose the non-smooth terrain into different convex and even patches, and rely on Mixed-Integer Programming to find the best patches for each footstep \cite{Deits2014,Cabezas2018}. However the computation time of those approaches make them difficult to use on a real robot. Overall, optimization techniques are not well suited for collisions and all of the presented approaches ignore this problem.

Another common approach is to rely on Graph Search \cite{Kuffner2001,Perrin2013,Winkler2015ICRA}. The space of possible footsteps is discrete and actions are selected using graph search algorithms, such as A*. However, such approaches quickly become too computationally expensive when solving for a large number of footsteps and/or considering the movement of the main body.

A final set of methods are based on machine learning.
Approaches based on supervised learning \cite{Cabezas2018,Holden2017} take the foosteps generated by a planner, or from motion capture as an input, so the resulting plan will be efficient/feasible only if the initial planner or the motions captured correspond to the capabilities of the robot.
Reinforcement learning has shown more and more impressive results during the last few years \cite{deepmind,Hwangbo2019} but, as of yet, the results are limited to either flat ground or to behaviors that are unsuitable for real robot hardware on other terrains.  

In this paper, we apply the work of Steve Tonneau et al. \cite{Tonneau2018} on our robotic platform, ANYmal \cite{anymal2016}, and explain some of the adjustments that need to be done to successfully compute feasible contact plans.

The next Section describes in more detail the decomposition of the contact planning problem and the different algorithms used to solve it.
Section \ref{sec:adjustment} discusses the different adaptations and tuning that were necessary to compute more realistic contact plans and Section \ref{sec:results} shows the resulting contact plans and the tests of those trajectories in a physically realistic simulation environment.

\section{Planner Description} \label{sec:planner}

\begin{figure}[tb!]
\centering
  \includegraphics[width=\linewidth]{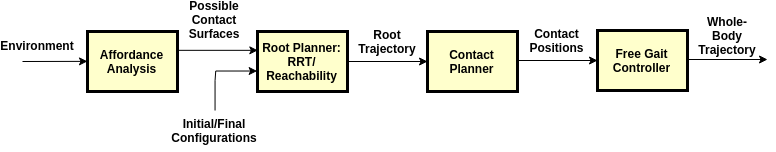}
  \caption{The steps of the acyclic reachability-based planner.}
  \label{fig:structure}
\end{figure}

Fig. \ref{fig:structure} shows the general structure of the pipeline used to generate a whole-body trajectory.
First, an algorithm analyses the environment to extract the set of possible contact surfaces. The planning problem is then decomposed into two subproblems, as described in \cite{Tonneau2018}; the algorithm searches for a trajectory of the main body of the robot, then contacts are created along this trajectory. This decomposition allows for a considerable reduction in problem complexity, as after the trajectory for the main body is found each limb is considered separately. The following Sections explain these different blocks in more detail.

\subsection{Foothold Affordances}
As a first step, the algorithm analyses the entire environment model to find which surfaces can be used to generate contacts. In this case, we consider the surfaces on which the robot can push. The criteria used to select whether a surface can be a contact surface are its inclination with respect to the vertical axis and the minimum size of the affordance.

Moreover, affordance analysis is used to avoid selecting contact points too close to an edge, to avoid the foot slipping and falling. Fig. \ref{fig:affordance} shows an example of possible contact surfaces after such affordance analysis.

\subsection{Root Planner}
A guide path for the main body of the robot is generated, in which static equilibrium is feasible. Some previous works have sampled for static equilibrium’s feasibility at intervals along the guide path \cite{Bouyarmane2009}.
However this is a very taxing process, so equilibrium feasibility is in this part approximated by contact reachability. This maintains the low problem  dimensionality and minimises computation time. 

\begin{figure}[tb!]
  \centering
  \begin{subfigure}[t]{0.28\linewidth}
    \includegraphics[width=\linewidth, trim={50 100 50 100}, clip]{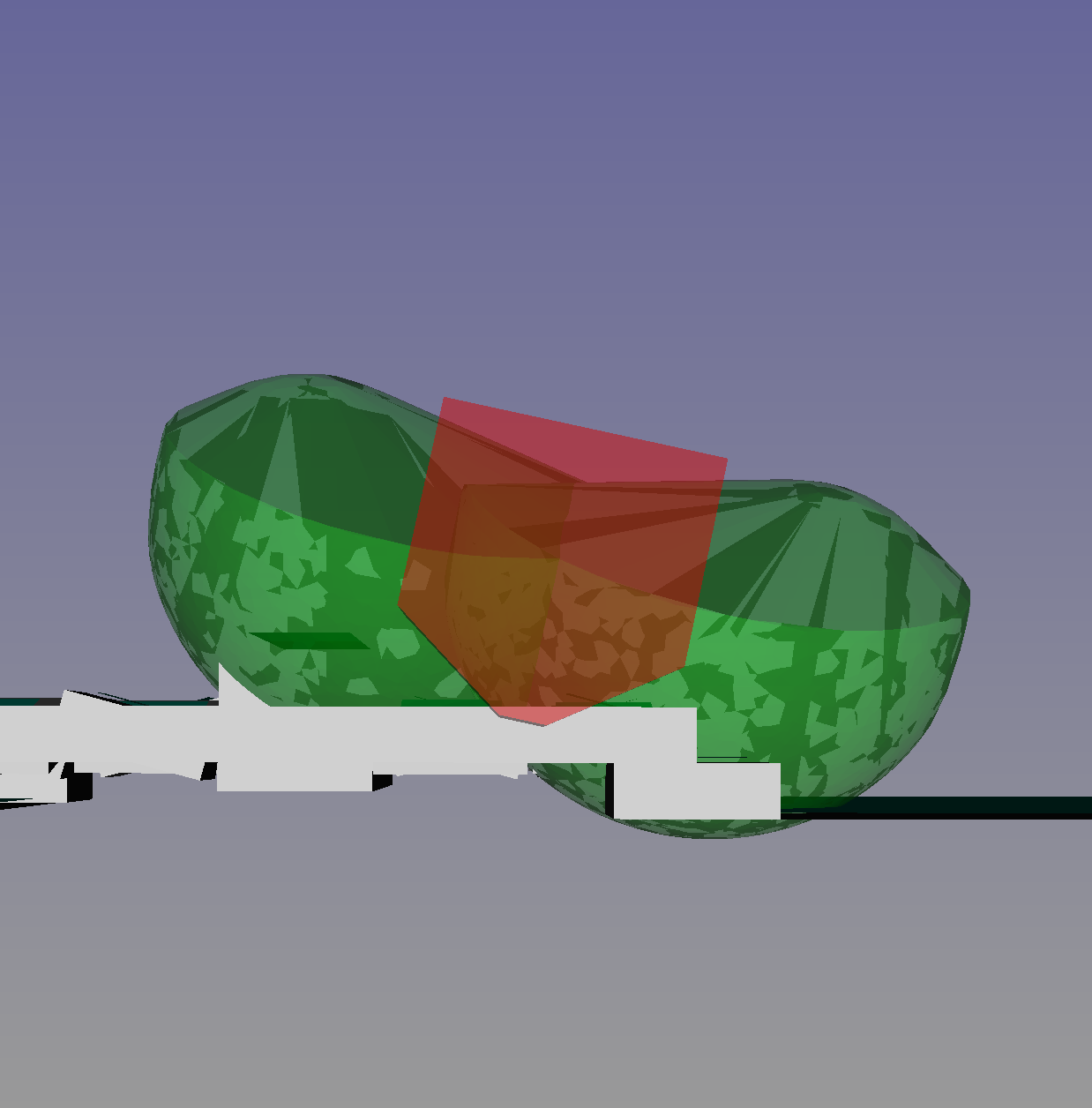}
    \caption{Risk of collision.}
  \end{subfigure}
  \begin{subfigure}[t]{0.28\linewidth}
    \includegraphics[width=\linewidth, trim={50 100 50 100}, clip]{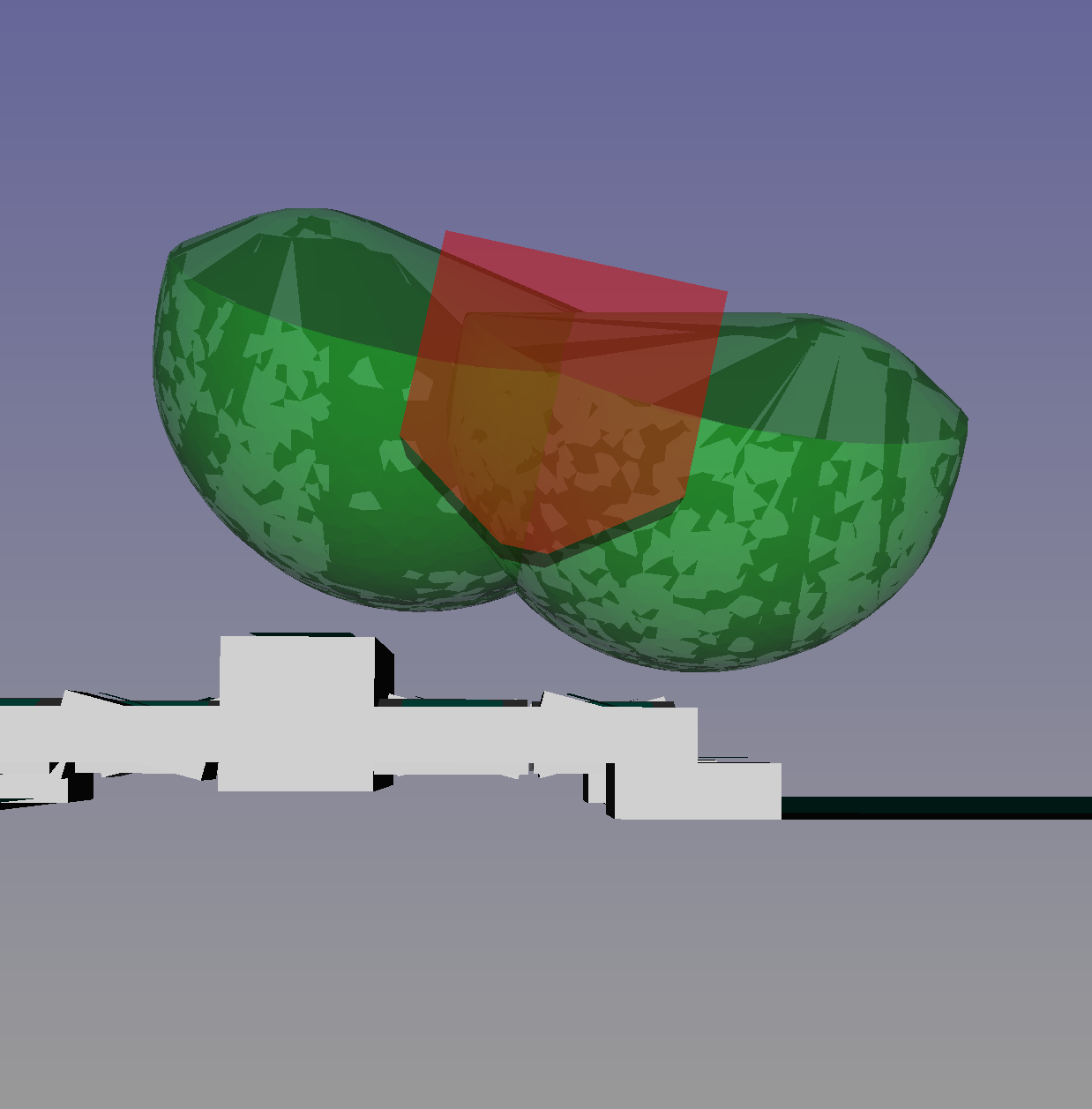}
    \caption{Not reachable.}
  \end{subfigure}
  \begin{subfigure}[t]{0.28\linewidth}
    \includegraphics[width=\linewidth, trim={50 100 50 100}, clip]{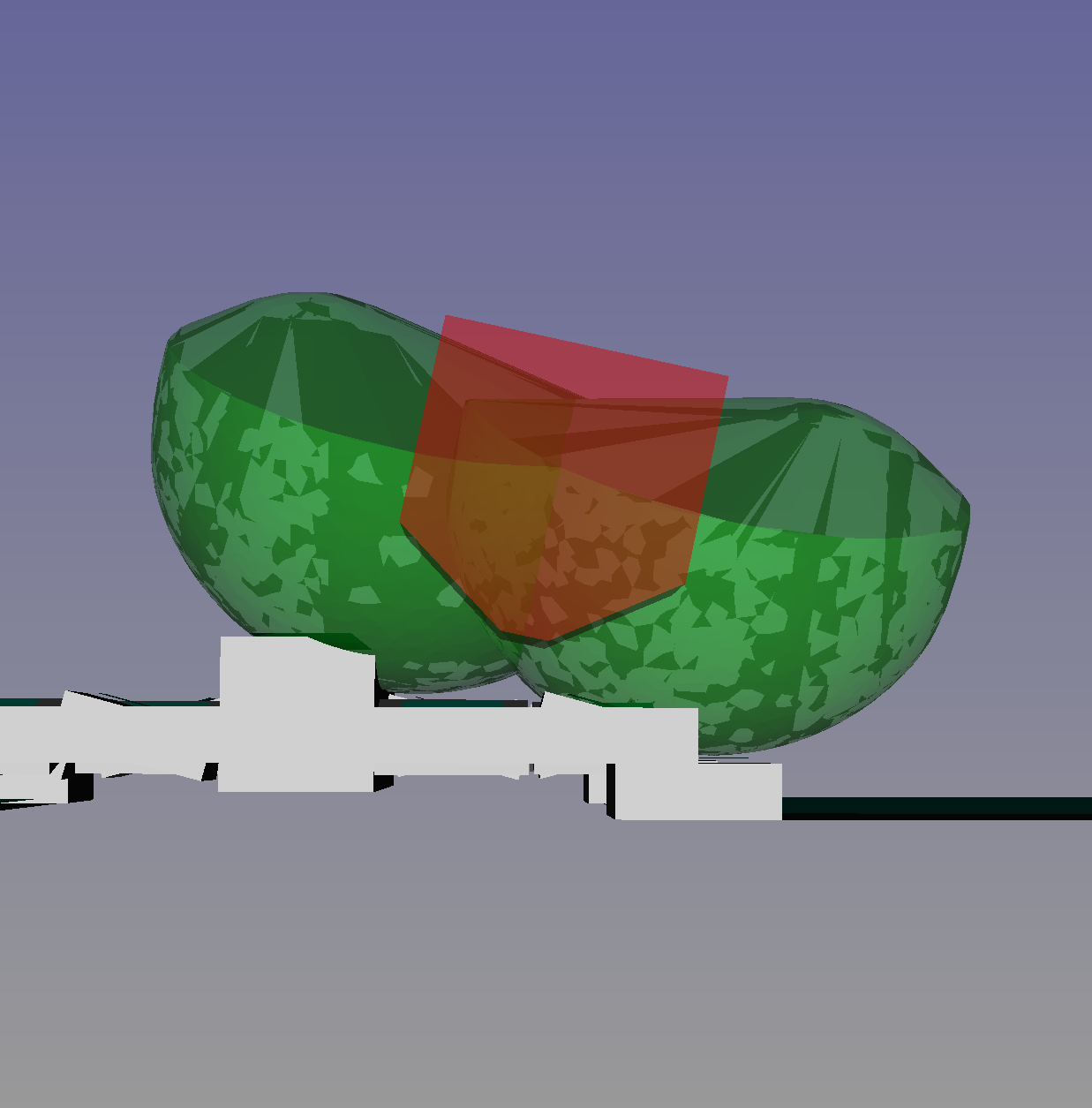}
    \caption{Valid body position.}
  \end{subfigure}
  \caption{The reachability condition is met by the figure on the right, and not by the other two. The root (red) is free from colliding with the environment geometry while the reachable space of the limbs (green) intersect the environment, meaning that contact can be created.}
  \label{fig:reachabilityCriterion}
\end{figure}

A root configuration is said to be \textit{contact reachable} if the environment intersects with the limb workspace and not the main body. If the main body intersects the environment, this implies collision, but if the environment does not intersect with the limb workspace then the robot cannot reach the environment to create contact. Therefore the region between these extremes, in which contact can be created without the body colliding, is considered to meet the reachability condition.
Fig. \ref{fig:reachabilityCriterion} shows several examples where only the last one is considered a valid body position.

A root path is then planned using an optimised Bi-RRT algorithm, propagating a random tree from both the start and goal positions to rapidly generate a complete trajectory, with sub-trajectories being validated by the reachability condition.
In addition, planing of the root trajectory is done in both position and velocity spaces, i.e.\ kinodynamic planing, as explained in \cite{Fernbach2017}.

\subsection{Contact Planner}
Given an initial whole-body configuration and a root trajectory, a sequence of whole-body configurations following this trajectory is computed, each separated by one step, to finish at the goal configuration. A step is defined as the breaking of one contact with the environment, followed by the creation of another contact for the same limb. 
This means that for each configuration, all 4 legs are in contact, which in turn means the current approach is limited to walking gaits, that can potentially be acyclic.

\subsubsection{Selection of the Stepping leg.}
The root trajectory is first discretized into equidistant intervals. On each interval, an inverse kinematic algorithm moves the root of the robot while trying to maintain the contacts. If it succeeds without collision, all contacts are maintained. If one contact cannot be maintained, the corresponding leg will be the stepping leg. If several contacts need to be broken, the interval is further subdivided so that only one contact will have to change.

On flat terrains, this algorithm will naturally result in a cyclic gait. However on more difficult terrains, where the footsteps can have different lengths or parts of the environment interfere with the movements of the robot, the algorithm is able to adapt and generate acyclic motions.

\begin{figure}[tb!]
  \centering
  \begin{subfigure}[h!]{0.4\linewidth}
    \includegraphics[width=\linewidth, trim={0 200 0 300}, clip]{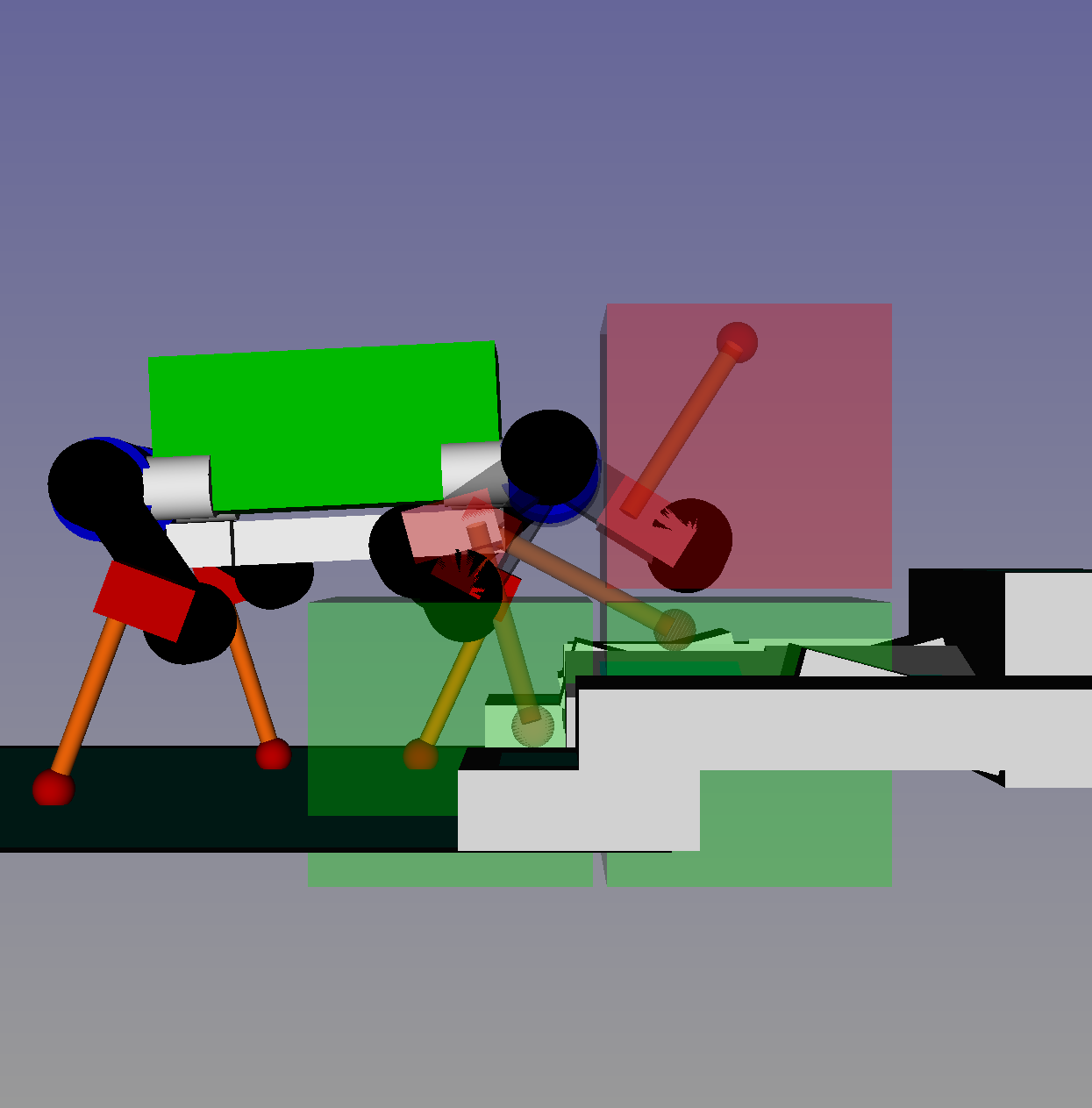}
    \caption{The octree and the environment intersect.}
  \end{subfigure}
  \begin{subfigure}[h!]{0.4\linewidth}
    \includegraphics[width=\linewidth, trim={0 200 0 300}, clip]{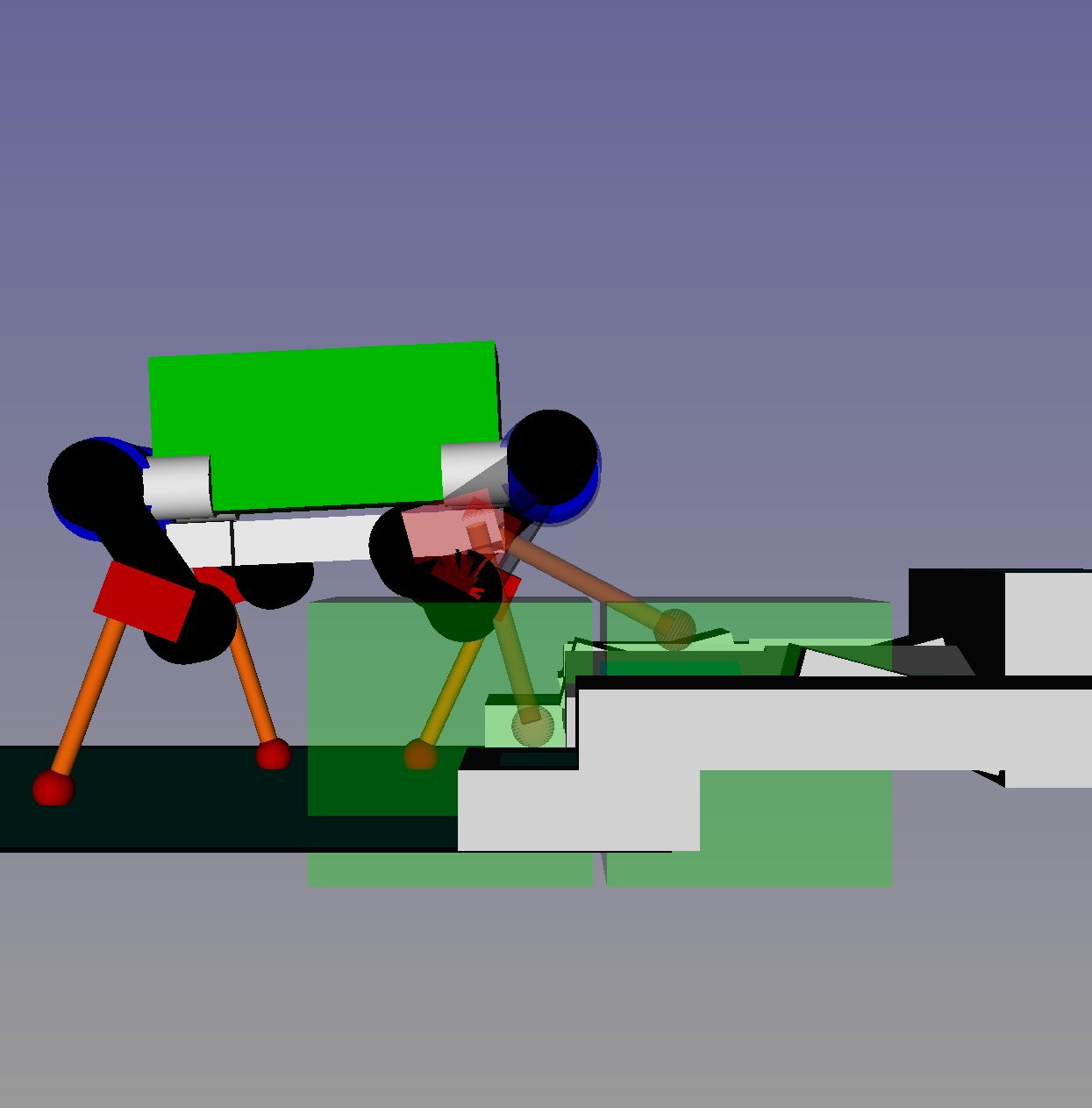}
    \caption{Keep only the configurations close to a surface.}
  \end{subfigure}
  \begin{subfigure}[h!]{0.4\linewidth}
    \includegraphics[width=\linewidth, trim={0 200 0 300}, clip]{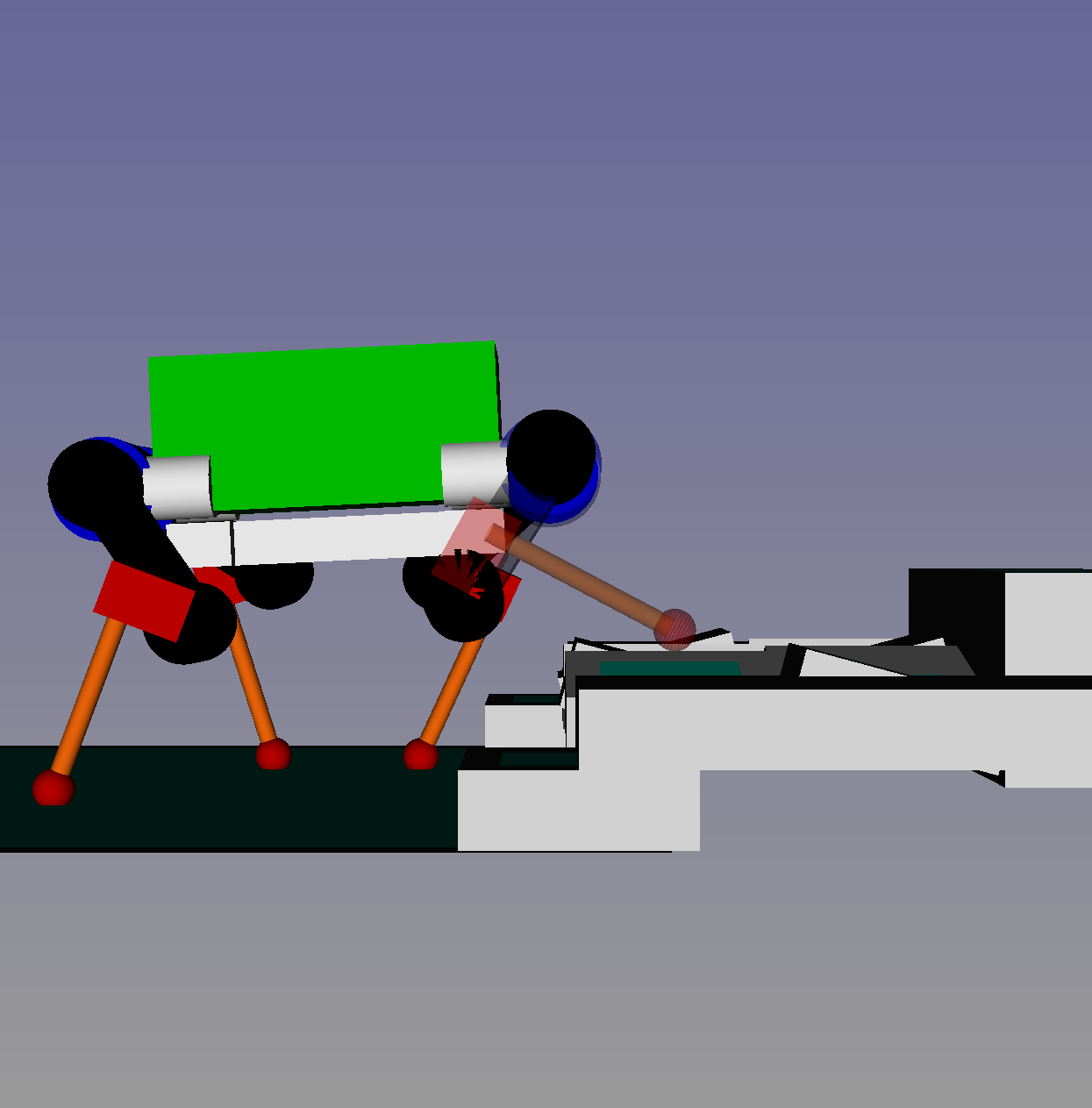}
    \caption{Select a configuration according to the heuristic.}
  \end{subfigure}
  \begin{subfigure}[h!]{0.4\linewidth}
    \includegraphics[width=\linewidth, trim={0 200 0 300}, clip]{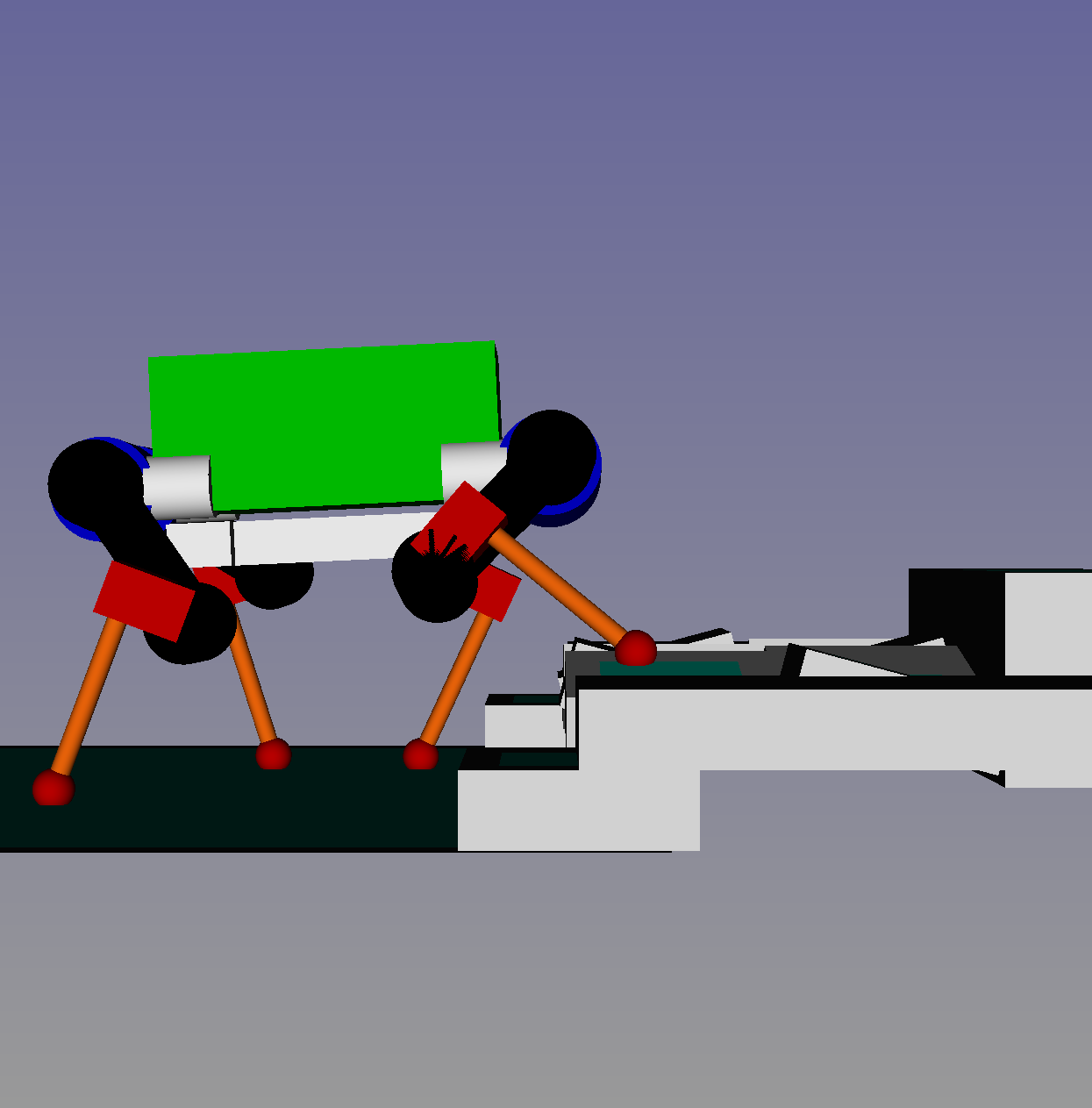}
    \caption{Project to a surface; check collisions and stability.} 
  \end{subfigure}
  \caption{Generation of a new contact.}
  \label{fig:contact_generation}
\end{figure}

\subsubsection{Contact Generation.}

Once the stepping leg is selected, the algorithm needs to project the corresponding foot to the new contact location. To reduce the computation time and select both the most suitable contact location and leg configuration, a database of leg configurations is used.

Offline, a database of random configurations is generated and stored in an octree data structure according to the foot position. Online, this octree is intersected with the environment to retrieve a set of leg configurations close to contact. Then, one of these configurations is selected (using a user-defined heuristic) to be projected into contact.

If the projection of the leg succeeds and the resulting whole-body configuration is statically stable and without collision, the configuration is kept and the algorithm continues to the next step. If not, the next (according to the heuristic) leg configurations are tested until a valid whole-body configuration is found.

Moreover, we check that there exist a feasible dynamic transition between each configuration using the algorithm presented in \cite{Fernbach2018}.
An example of the resulting sequence of configuration is shown in Fig. \ref{fig:contact_generation}.


\section{Adjustments to ANYmal} \label{sec:adjustment}

The algorithm is able to generate contact plans for any robot morphology on different terrain, nonetheless it relies on different 
user inputs that need to be adjusted for the robot at hand.
These 
user inputs are mainly the shapes (range of motions and non-collision with the main body) and the heuristics used to select between leg configurations in the octree. The next sections present how these 
inputs have been adjusted to generate more realistic contact plans for the ANYmal quadruped robot.

\subsection{Adjusting the Shapes for the Root Planner}

The ranges of motion for each limb are generated by sampling random configurations. Only configurations in the range of the motor are sampled and configurations that result in self-collision are rejected. The position of the feet are saved, then ROMs are generated by constructing the convex hull of those foot positions.

\begin{figure}[tb!]
  \centering
  \begin{subfigure}[h!]{0.28\linewidth}
    \includegraphics[width=\linewidth, trim={0 200 0 0}, clip]{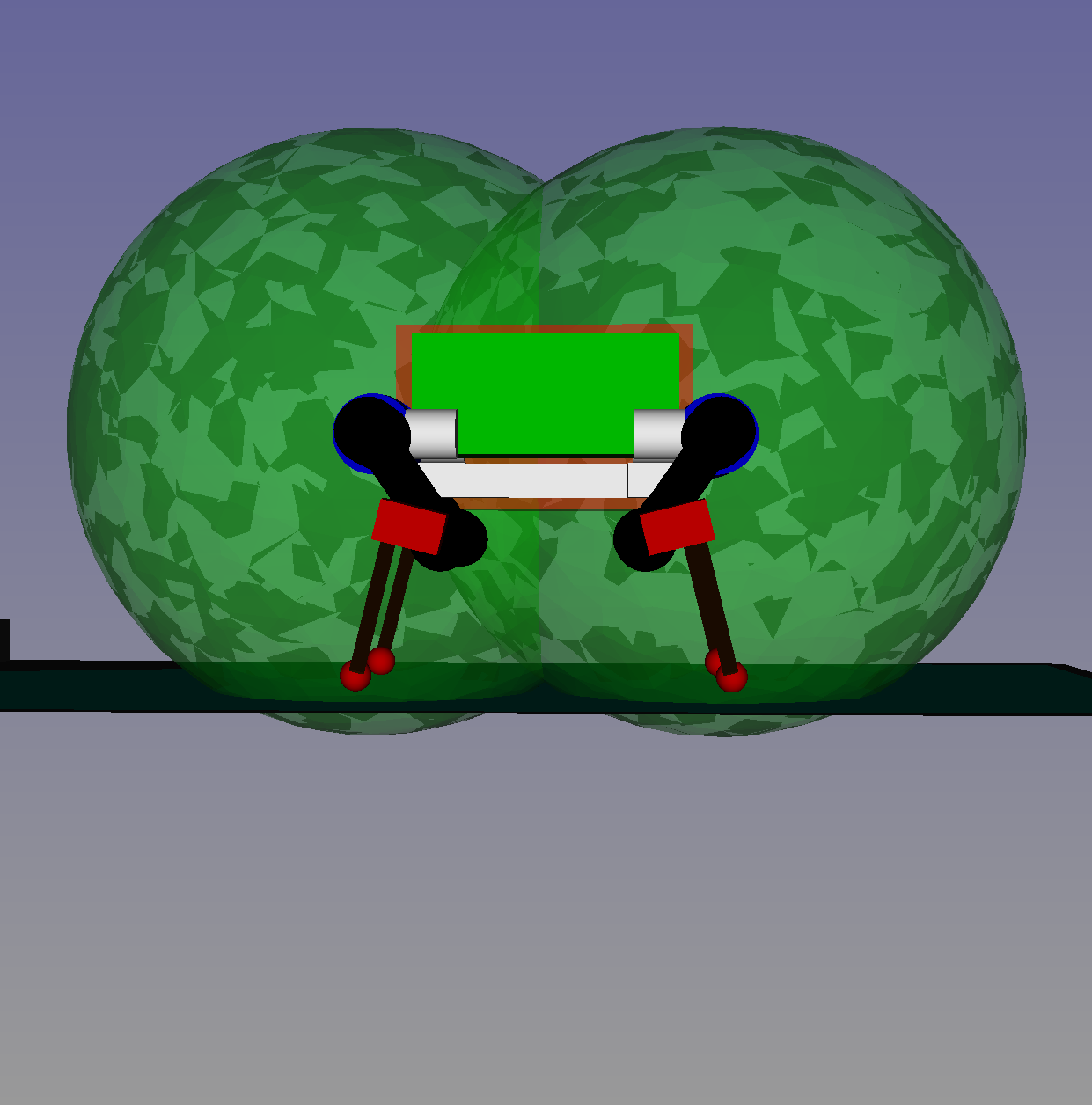}
    \caption{Reachability.}
  \end{subfigure}
  \begin{subfigure}[h!]{0.28\linewidth}
    \includegraphics[width=\linewidth, trim={0 200 0 0}, clip]{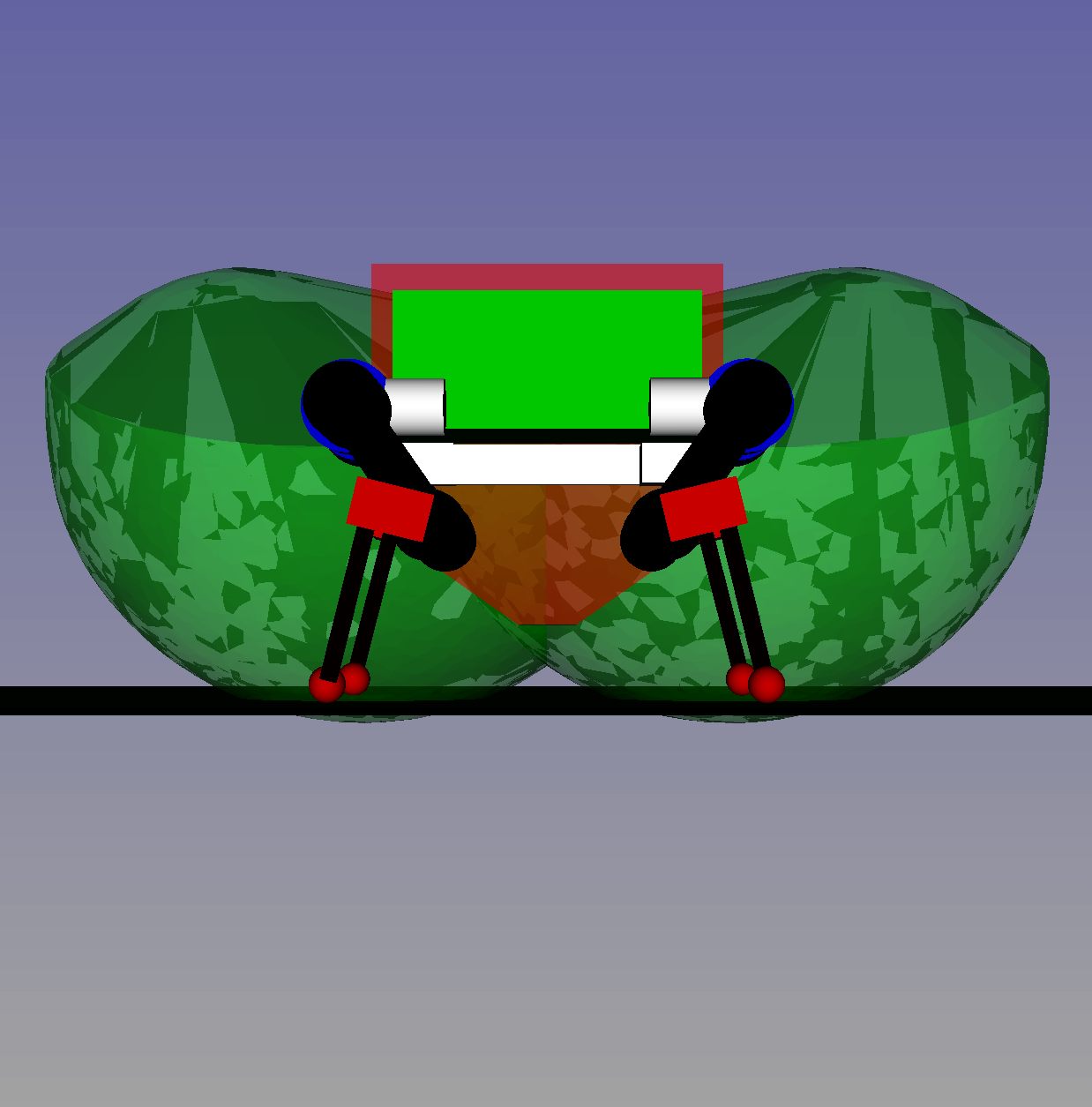}
    \caption{Tuned Reachability.}
  \end{subfigure}
  \begin{subfigure}[h!]{0.28\linewidth}
    \includegraphics[width=\linewidth, trim={0 100 0 100}, clip]{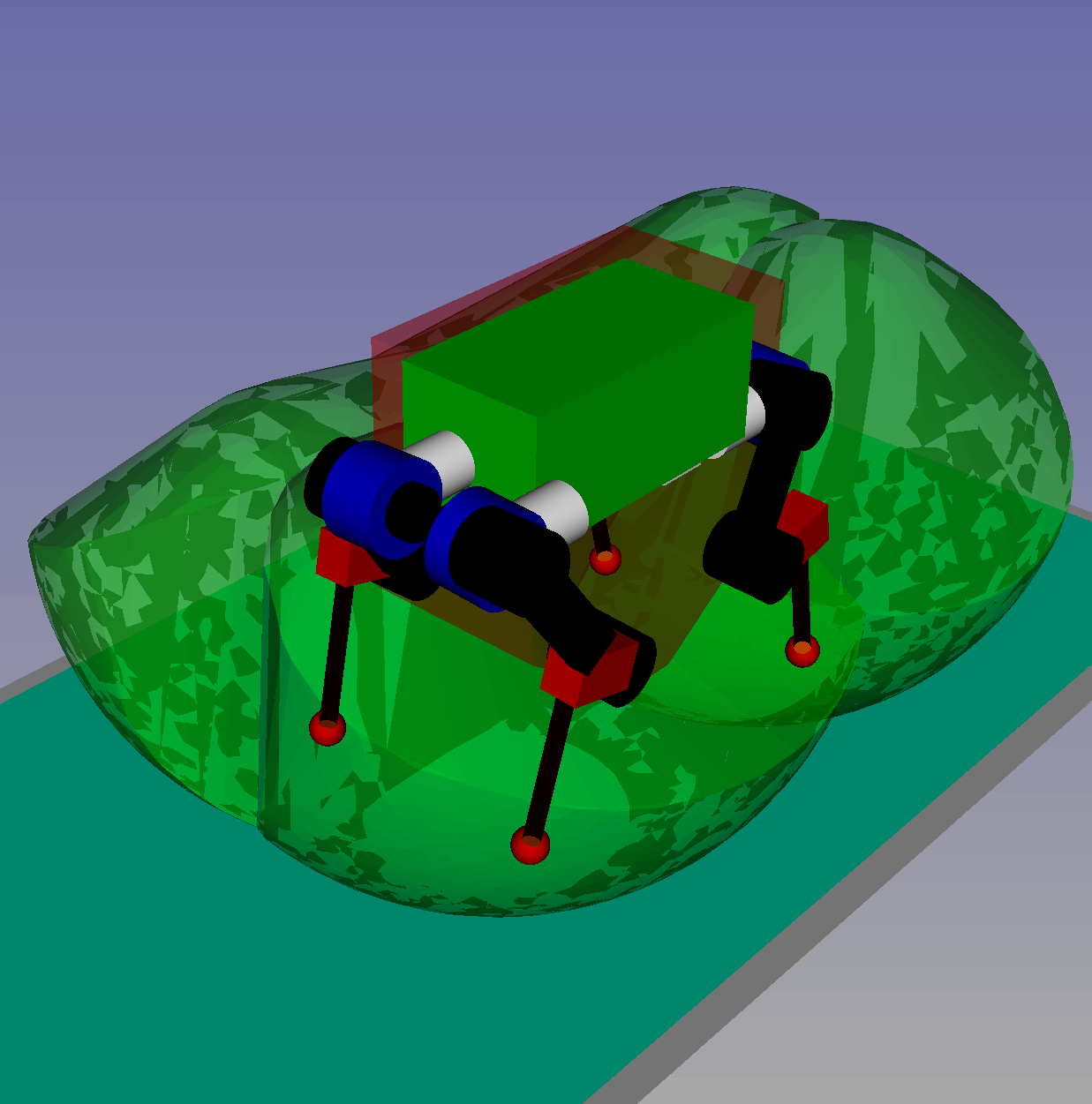}
    \caption{Tuned Reachability.}
  \end{subfigure}
  \caption{The range of motion of the ANYmal robot before (left) and after (centre and right) retuning.}
  \label{fig:reachabilityMesh}
\end{figure}

While this approach is valid for robots where the range of motion is limited enough to actually correspond to possible contact positions, ANYmal's joints allow \ang{360} rotations in both directions. 
To avoid the robot attempting to walk upside-down (technically possible, but not suitable) and to avoid the robot walking with large steps that could generate high torques at the hips, the range of the joints are limited and the range of motion is reduced by a factor of $0.85$. Fig. \ref{fig:reachabilityMesh} shows the different range of motions, before and after retuning.

Moreover, if the main body is close to the ground, the torques in the legs become prohibitively large and the space where the legs can feasibly make contact without colliding with the main body is greatly reduced. Therefore, a shape corresponding to the non-collision constraint is added close to the ground, as shown in Fig. \ref{fig:reachabilityMesh}(b) and (c). This `V' shape is used to allow smooth trajectories on terrains like stairs.

\subsection{Heuristics for Selecting the Leg Configurations in the Octree}

Each sample of the leg configuration is scored based on two sets of heuristics. The first uses all available offline parameters, to perform as much calculation as possible ahead of time and reduce the required online computation.

In our case, this part is computed as a weighted distance (in configuration space) between the sample configuration and the standard standing configuration of the robot. This cost is used so the robot keeps a relatively constant configuration and avoids the motors making \ang{360} rotations or constantly switching between ``X'' and ``O'' configurations.

\begin{figure}[tb!]
  \centering
    \begin{subfigure}[h!]{0.32\linewidth}
    \includegraphics[width=\linewidth, trim={0 100 0 150}, clip]{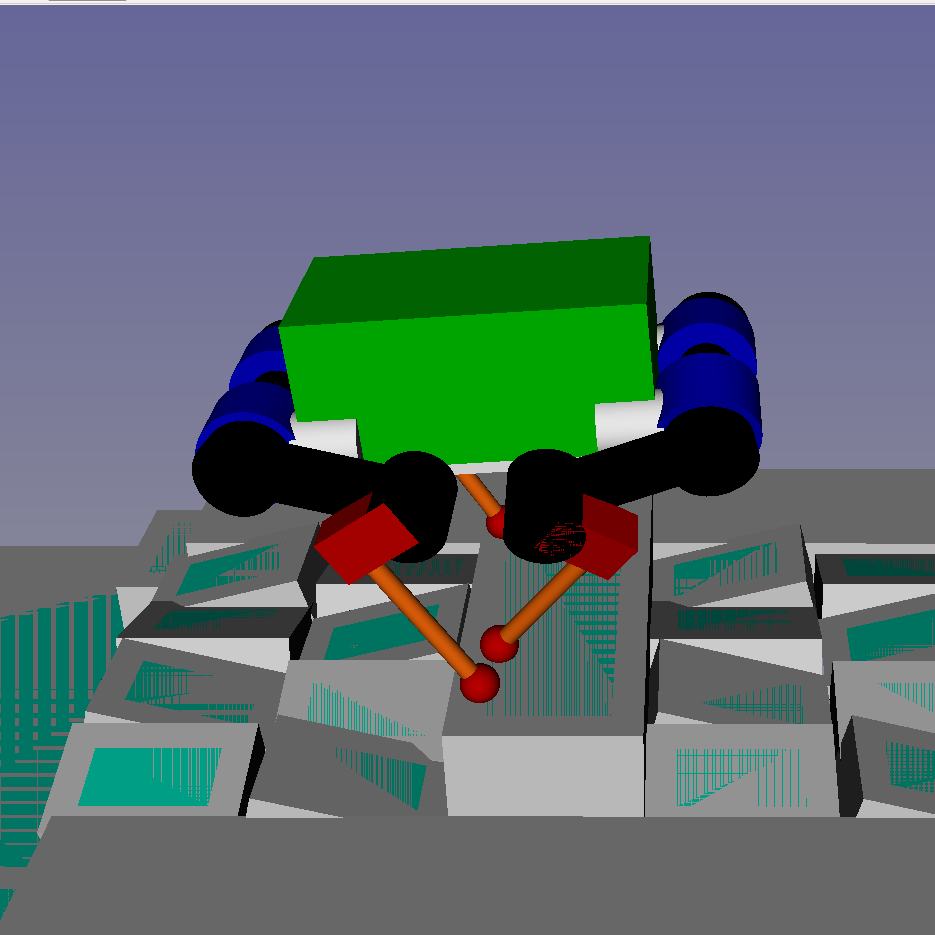}
    \caption{}
    \label{fig:heuristics1}
  \end{subfigure}
  \begin{subfigure}[h!]{0.32\linewidth}
    \includegraphics[width=\linewidth, trim={0 70 0 150}, clip]{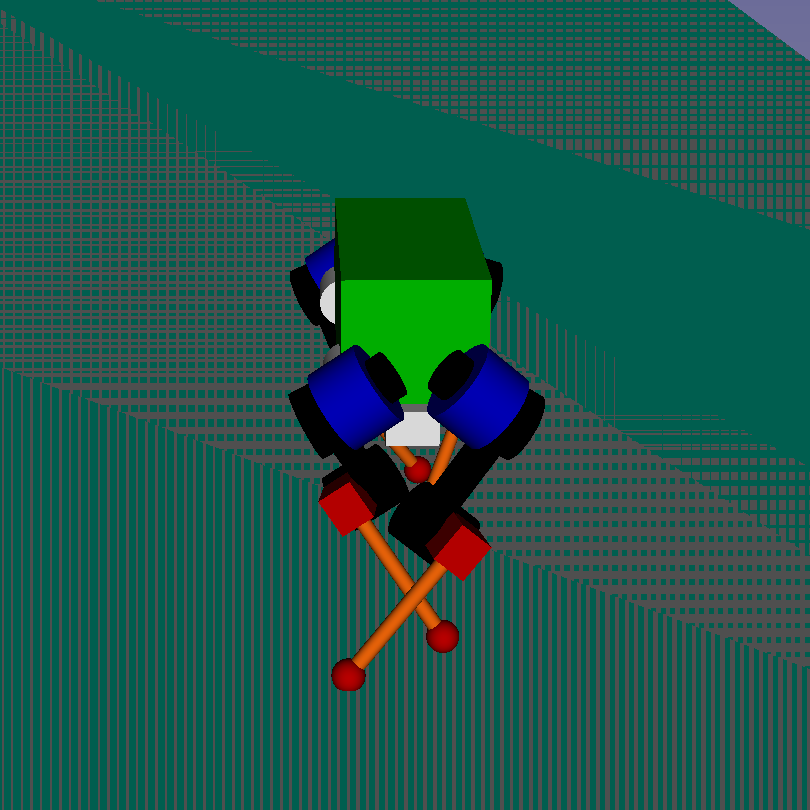}
	\caption{}    
    \label{fig:heuristics2}  
  \end{subfigure}
  \caption{Examples of instability due to poor heuristics. On the left, the hind legs' reference position is too far forward under a forward motion. On the right, the legs' reference position is too far over in the direction of lateral motion.}
  \label{fig:heuristics}
\end{figure}

The second set of heuristics uses parameters that can only be determined online, such as environment slope and the robot's direction of motion. Samples close to the reference position are favoured to increase controllability, stability and maintain motion towards the goal. 
This reference position is set as the position of the foot for a reference limb configuration and a main body position at time $t + \Delta t$. $\Delta t$ must then be adjusted to avoid the support polygon becoming too small as in Fig. \ref{fig:heuristics1} or to prevent limbs being placed far across from the body and overlapping other legs as in Fig. \ref{fig:heuristics2}. 

\section{Results} \label{sec:results}

This Section shows the results obtained with the contact planner and the experiments in physically realistic simulation, using the Gazebo simulator. All trajectories shown are obtained using the open source planner HPP and its implementation of the reachability-based planner \cite{hpp,hpp-link}.

For the set of weights, parameters and shapes used in this work please refer to
\url{https://github.com/Mathieu-Geisert/hpp-rbprm} and
\url{https://github.com/Mathieu-Geisert/hpp-rbprm-corba} under branch ``anymal.''

\begin{figure}[tb!]
  \centering
  \begin{subfigure}[h!]{0.4\linewidth}
    \includegraphics[width=\linewidth]{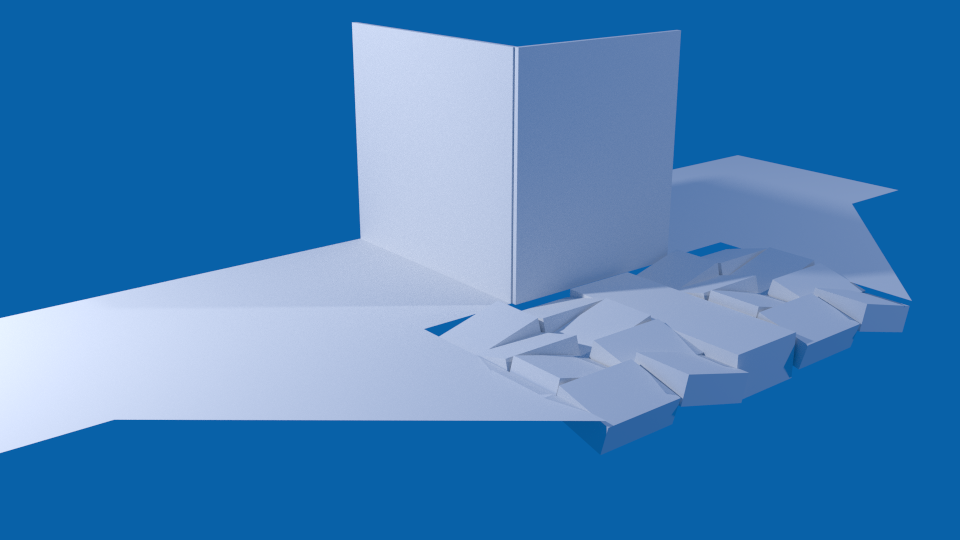}
    \caption{Slalom rubble.}
    \label{fig:env_slalom}
  \end{subfigure}
  \begin{subfigure}[h!]{0.4\linewidth}
    \includegraphics[width=\linewidth]{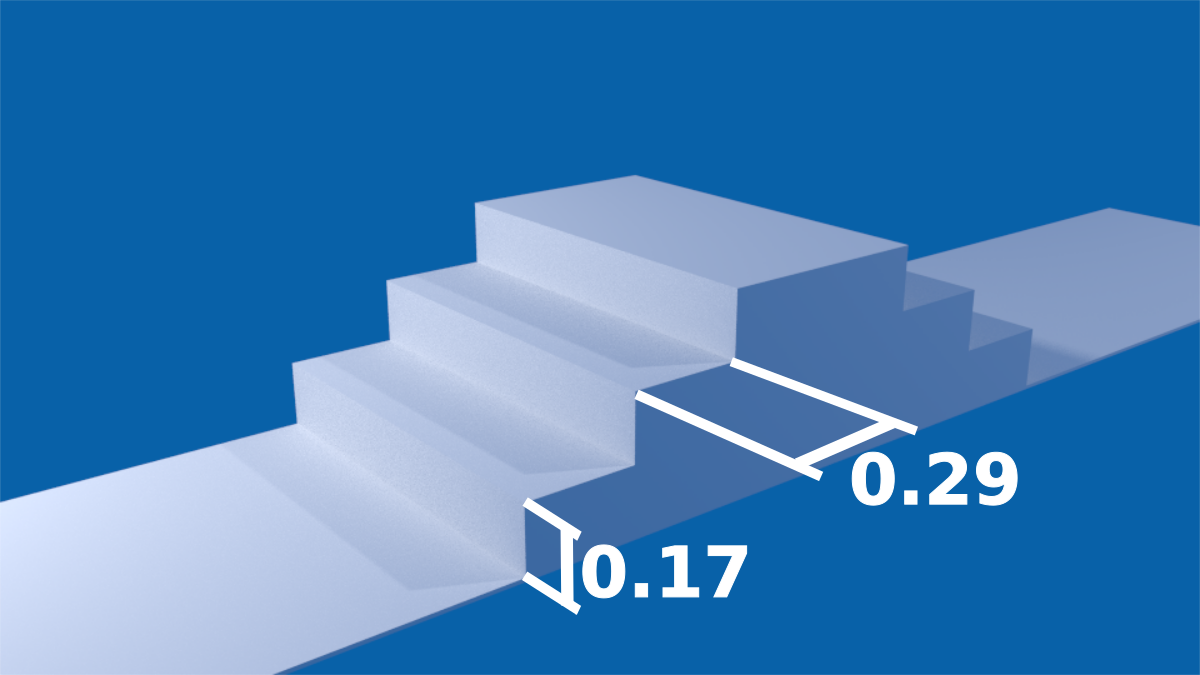}
    \caption{Plinth.}
    \label{fig:env_plinth}
  \end{subfigure}
  \vskip 2pt
  \begin{subfigure}[h!]{0.4\linewidth}
    \includegraphics[width=\linewidth]{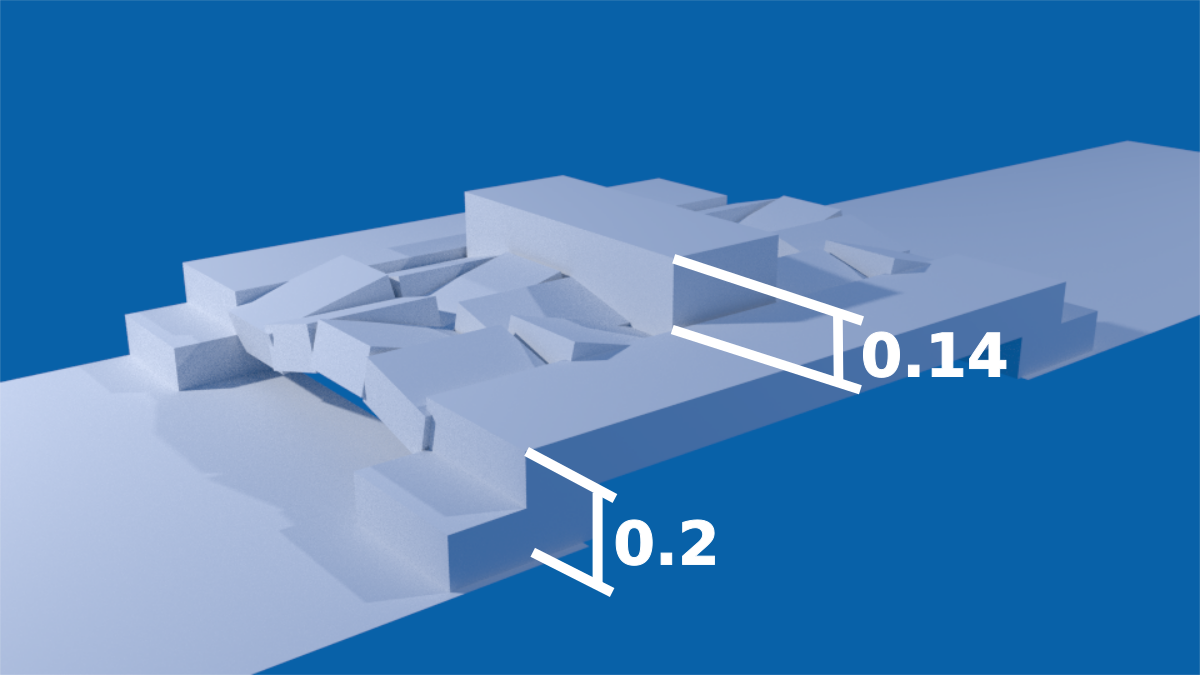}
    \caption{DARPA.}
    \label{fig:env_darpa}
  \end{subfigure}
  \begin{subfigure}[h!]{0.4\linewidth}
    \includegraphics[width=\linewidth, height=0.5625\linewidth, trim={0 0 0 0}, clip]{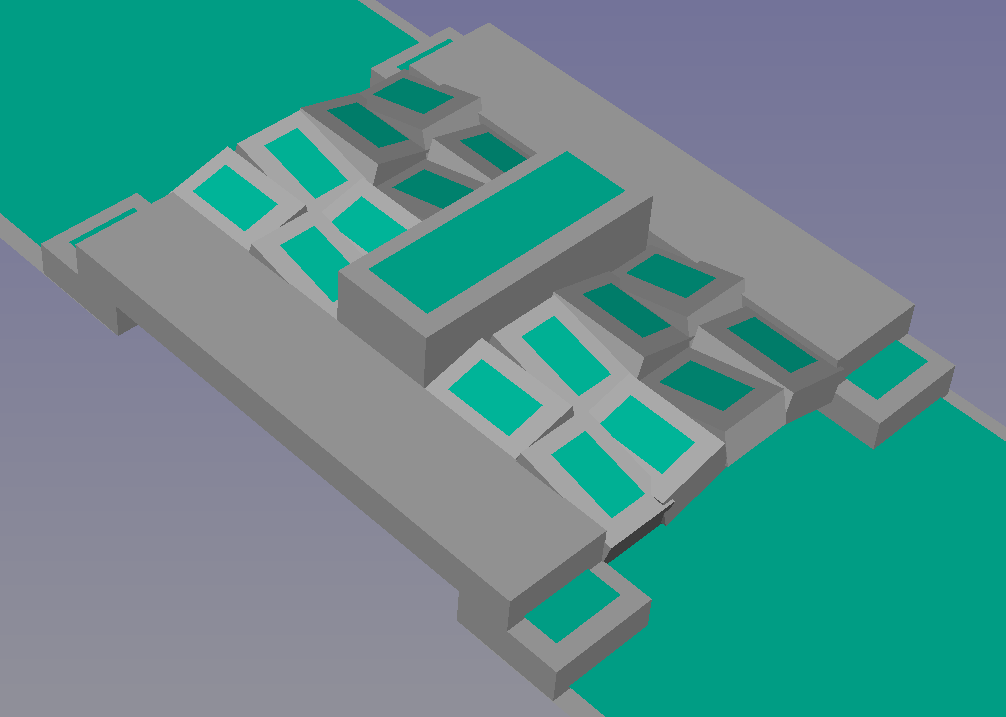}
    \caption{DARPA affordance.}
    \label{fig:affordance}
  \end{subfigure}
  \caption{Samples of test environments.}
  \label{fig:environments}
    \vskip 6pt
\end{figure}

The pipeline presented in this paper is tested on terrains of progressively varying difficulty:
\begin{itemize}
\item Flat floor.
\item Terrains with small height variation and obstacles like Fig. \ref{fig:env_slalom}.
\item Flat surfaces but with large height variation like the stairs in Fig. \ref{fig:env_plinth}.
\item Non-flat surfaces with large height variation like the rubble terrain from the DARPA  Robotics Challenge final in 2015 shown in Fig. \ref{fig:env_darpa}.
\end{itemize}

\subsection{Generation of the contact plan}

\begin{figure}[tb!]
  \centering
  \begin{subfigure}[h!]{0.80\linewidth}
    \includegraphics[width=\linewidth, trim={0 450 0 250}, clip]{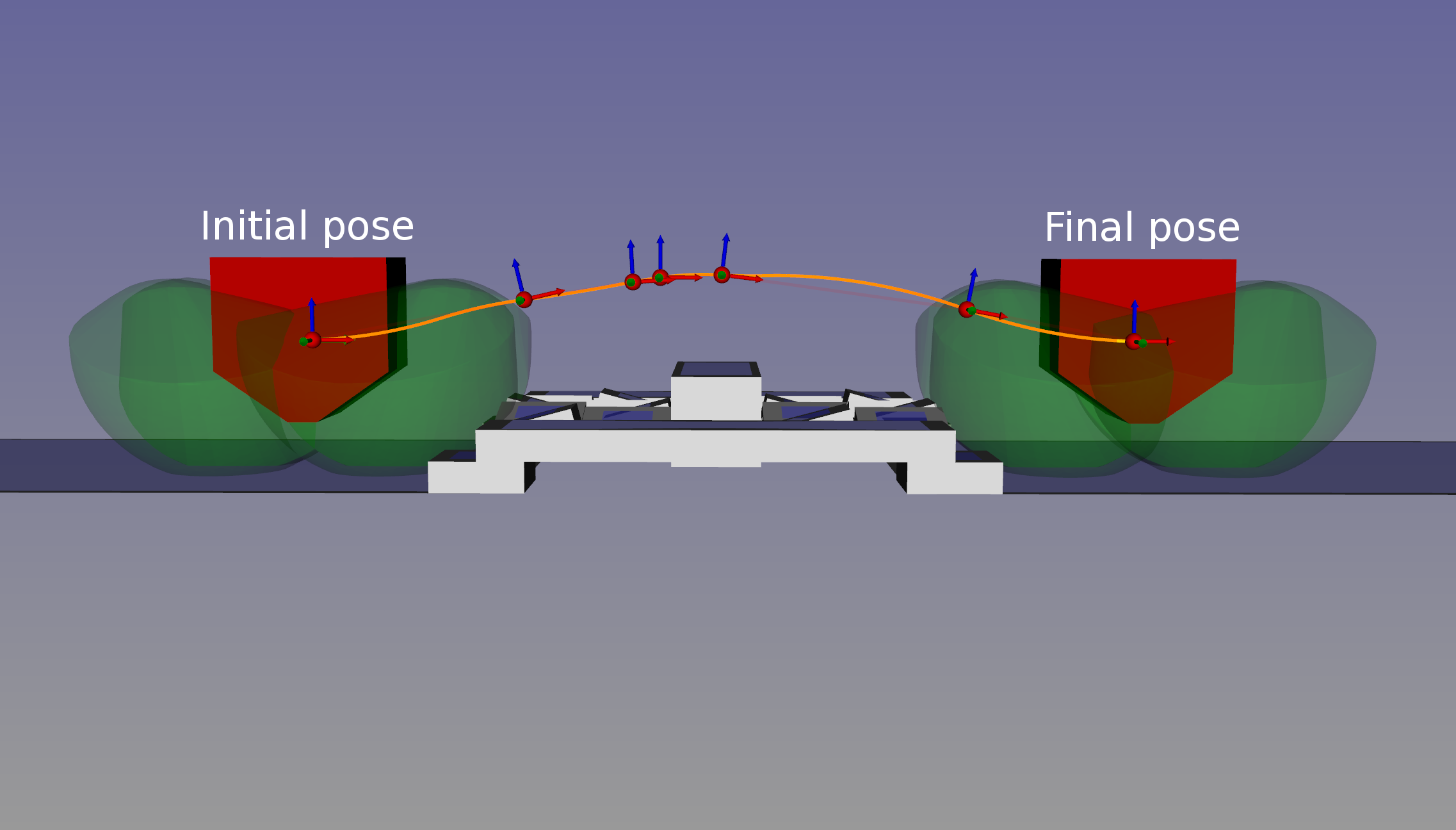}
  \end{subfigure}
    \vskip 2pt  
  \begin{subfigure}[h!]{0.26\linewidth}
    \includegraphics[width=\linewidth]{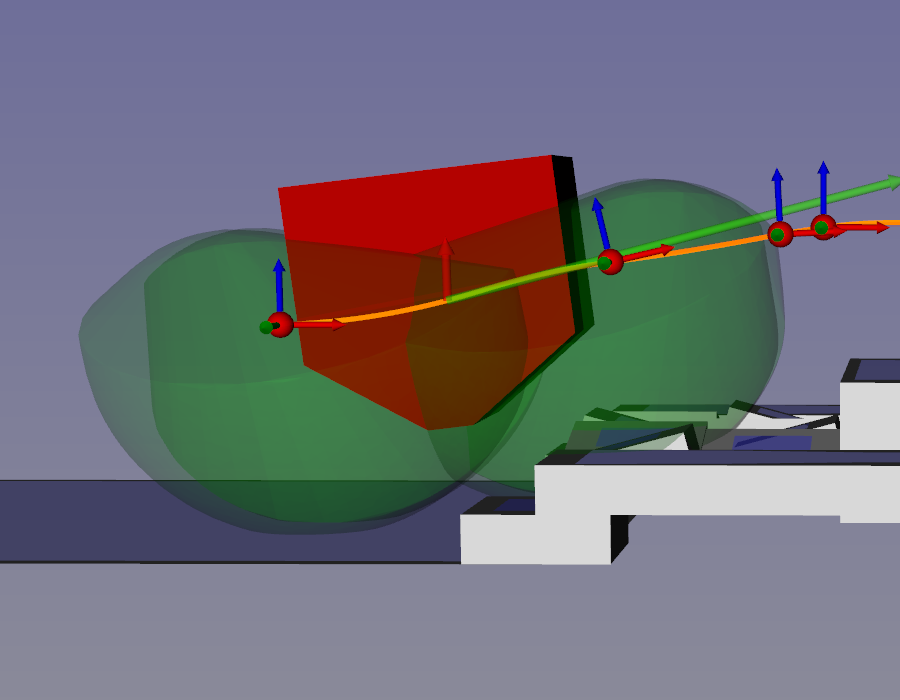}
  \end{subfigure}
  \begin{subfigure}[h!]{0.26\linewidth}
    \includegraphics[width=\linewidth]{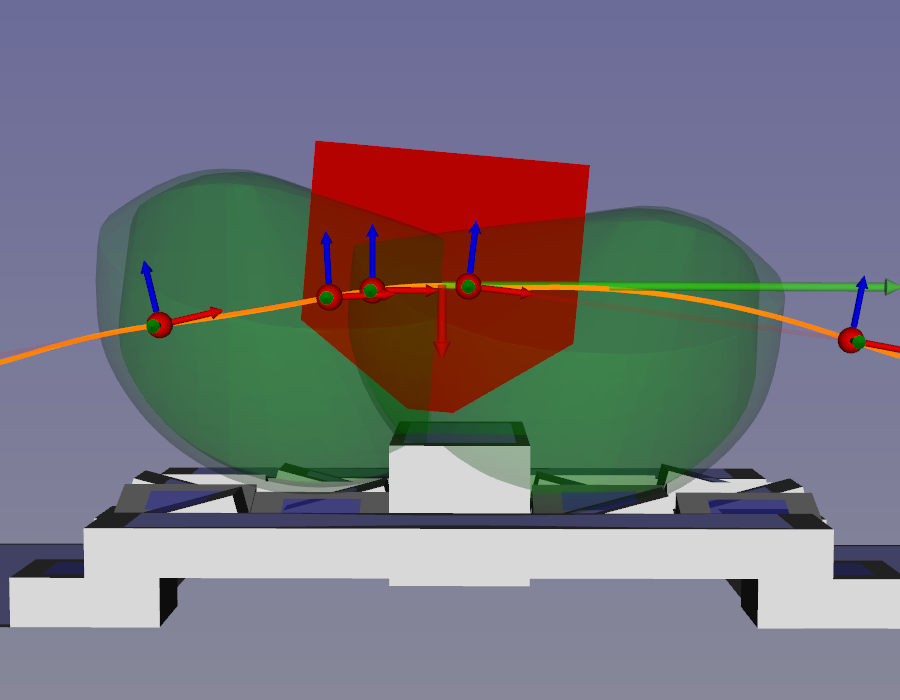}
  \end{subfigure}
  \begin{subfigure}[h!]{0.26\linewidth}
    \includegraphics[width=\linewidth]{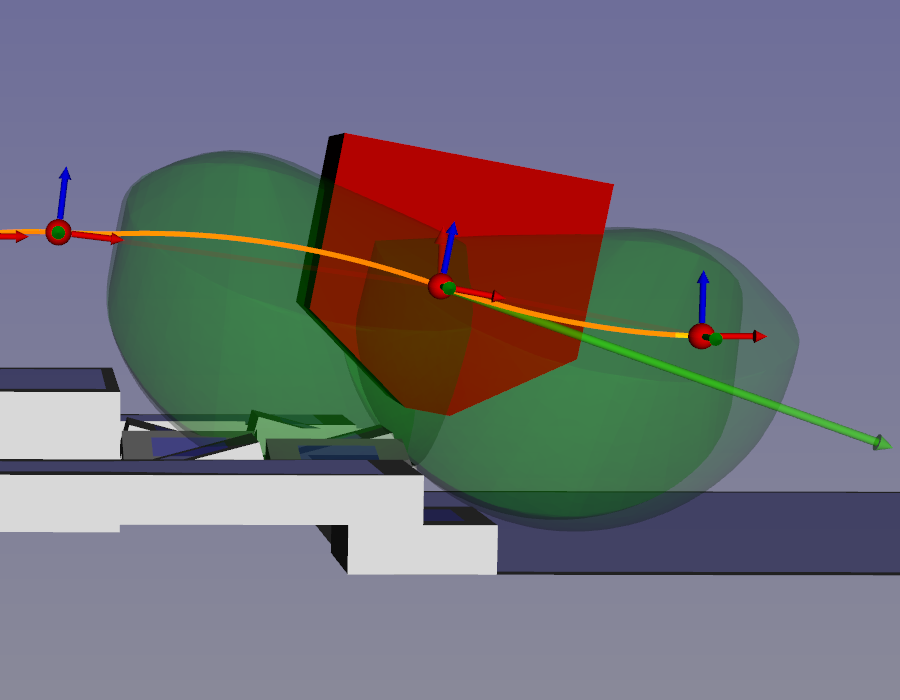}
  \end{subfigure}
  \caption{Top: the trajectory of the main body of the robot computed by the kinodynamic planner. Bottom: the reachability condition at several configurations.}
  \label{fig:trajectoryResult}
  
\end{figure}

The guide path planner has proven very robust, and has given trajectories for the root to follow in all tested environments, including the DARPA rubble terrain challenge.
In comparison with the previous version of the algorithm that was not based on kinodynamic planning, the generated trajectories are smooth and have no sudden change of direction. An example of root trajectory is shown in Fig. \ref{fig:trajectoryResult}. This smoothness allow us to use the motion of the root for the heuristic of the contact planner.


\begin{figure}[tb]
  \centering
   \begin{subfigure}[h!]{0.28\linewidth}
   \includegraphics[width=\linewidth]{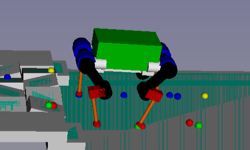}
  \end{subfigure}
   \begin{subfigure}[h!]{0.28\linewidth}
   \includegraphics[width=\linewidth]{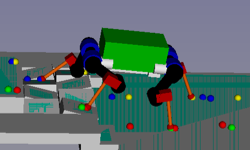}
  \end{subfigure}
   \begin{subfigure}[h!]{0.28\linewidth}
   \includegraphics[width=\linewidth]{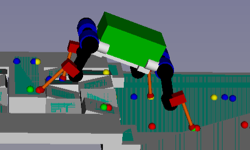}
  \end{subfigure}
  \vskip 2pt
     \begin{subfigure}[h!]{0.28\linewidth}
   \includegraphics[width=\linewidth]{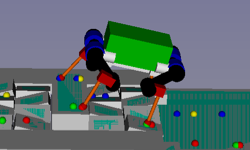}
  \end{subfigure}
     \begin{subfigure}[h!]{0.28\linewidth}
   \includegraphics[width=\linewidth]{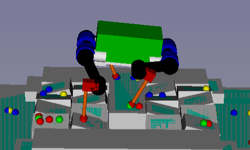}
  \end{subfigure}
     \begin{subfigure}[h!]{0.28\linewidth}
   \includegraphics[width=\linewidth]{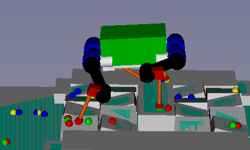}
  \end{subfigure}
  \vskip 2pt
     \begin{subfigure}[h!]{0.28\linewidth}
   \includegraphics[width=\linewidth]{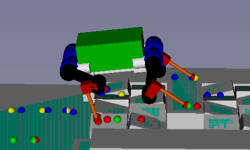}
  \end{subfigure}
     \begin{subfigure}[h!]{0.28\linewidth}
   \includegraphics[width=\linewidth]{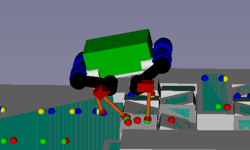}
  \end{subfigure}
     \begin{subfigure}[h!]{0.28\linewidth}
   \includegraphics[width=\linewidth]{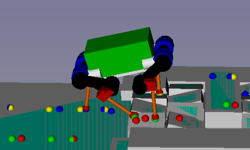}
  \end{subfigure}
  \vskip 2pt
     \begin{subfigure}[h!]{0.28\linewidth}
   \includegraphics[width=\linewidth]{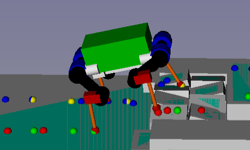}
  \end{subfigure}
     \begin{subfigure}[h!]{0.28\linewidth}
   \includegraphics[width=\linewidth]{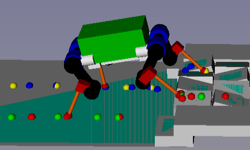}
  \end{subfigure}
     \begin{subfigure}[h!]{0.28\linewidth}
   \includegraphics[width=\linewidth]{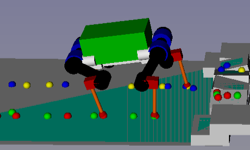}
  \end{subfigure}
  \caption{Contact plan over the DARPA environment, top to bottom and left to right, ANYmal is walking to the left. Each coloured ball represents a planned footstep, where each colour corresponds to a different leg.}
    \label{fig:contactPlanner}
\end{figure}

The contact planner implemented has been very effective at producing a set of footsteps in which static equilibrium is feasible. For each environment yet tested, the computation time can vary between each test but the planner always produces contacts that follow the guide path while avoiding collision and statically unstable configuration. This suggests that the reachability condition is a reasonable approximation to have static equilibrium for a quadruped robot. Fig. \ref{fig:contactPlanner} shows an example of sequence of configuration generated by the planner.

Tab. \ref{tab:computation_time} shows the computation time for each algorithm of the planner. For those tests, an important factor that influence the computation time is the size of the boxes of the octree and the number of generated samples for each limb. In those examples, we use \SI{1}{\cubic\centi\metre} boxes populated with 50000 samples. The generation of this octree takes about \SI{5.4}{\second}, but for an real usage on the robot, this octree would only need to be constructed once.


\subsection{Tests in Simulation}

As shown on Fig. \ref{fig:structure}, once a contact plan is generated we can use the Free Gait controller \cite{freegait} of the ANYmal to generate the corresponding trajectory for the whole-body. 
Trajectories are simulated in the Gazebo simulator, which includes constraints such as contacts, velocity and torque limits, that have not been explicitly addressed thus far in the planning pipeline.

Table \ref{tab:computation_time} shows the rate of success for the ANYmal to execute the contact plan. The rate of success is high on simple terrains -- and most failures come from the fact the robot does not take into account the environment when computing the leg trajectories -- but it quickly decreases on more complex terrains. 

An important part of the failures comes from too high torques. Even if the shape of the range of motion was scaled down in the root planner, the contact planner can keep a contacts until it becomes unreachable. However, in the environment with height variation, the torque limits are reached much sooner than the reachability limits.

Another part of the failures comes from a too small support polygon. Although the algorithm used to check stability is able to give us a robustness score, this score reflects the margins between the contact forces and their friction cones and not the margin with respect to the support polygon.


\begin{figure}[tb]
  \centering
   \begin{subfigure}[h!]{0.35\linewidth}
   \includegraphics[width=\linewidth, trim={250 0 150 100}, clip]{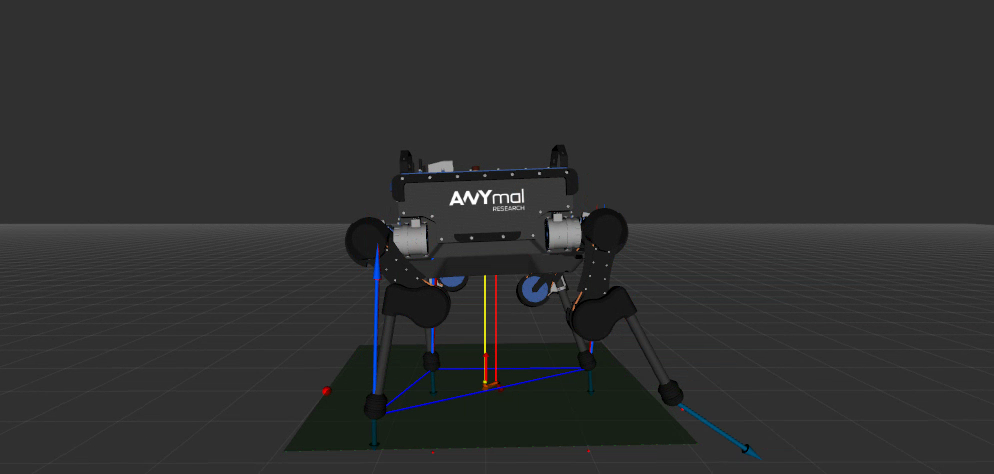}
  \end{subfigure}
   \begin{subfigure}[h!]{0.35\linewidth}
   \includegraphics[width=\linewidth, trim={500 400 700 235}, clip]{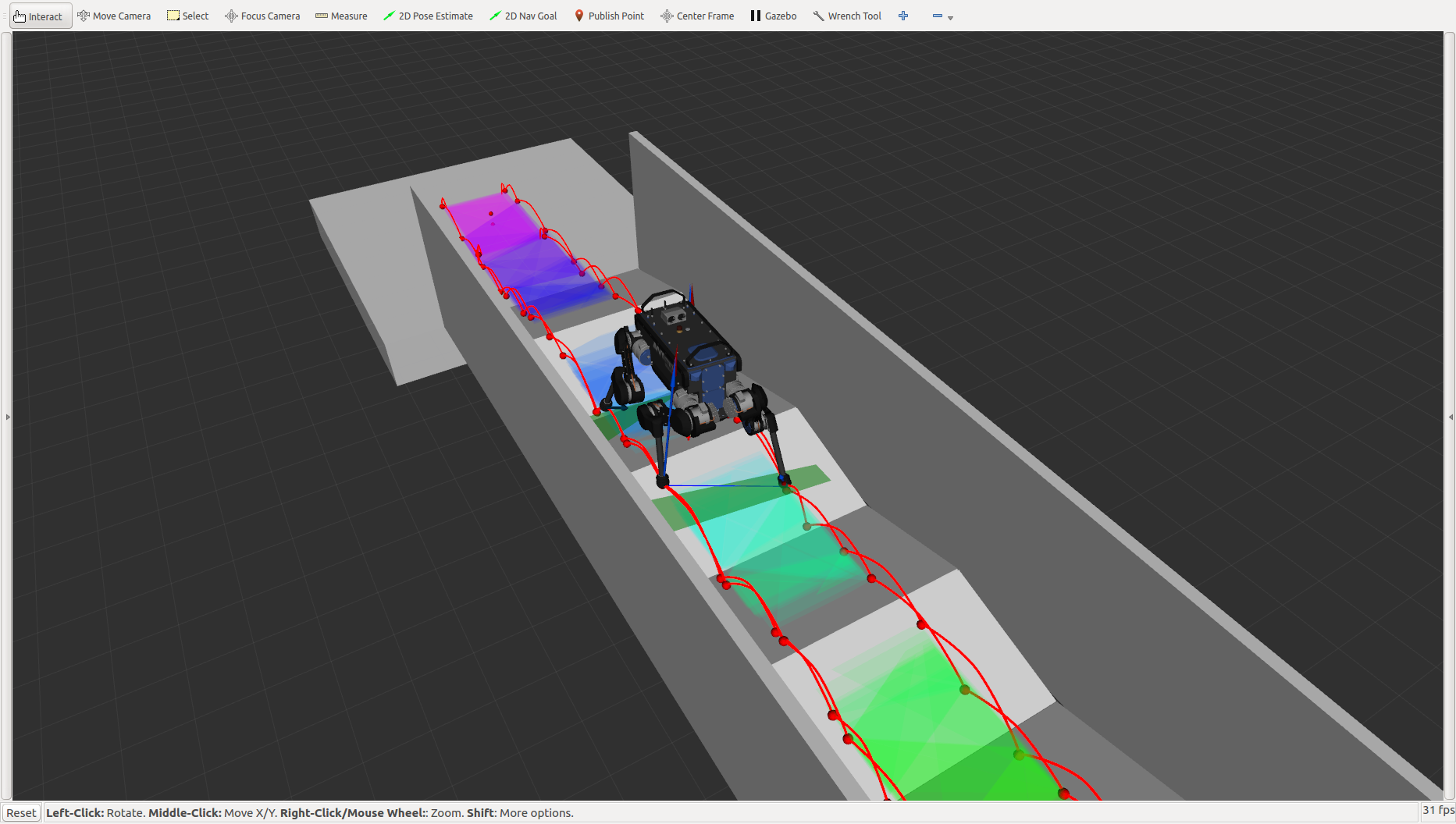}
  \end{subfigure}
  \vskip 3pt
   \begin{subfigure}[h!]{0.35\linewidth}
   \includegraphics[width=\linewidth, trim={520 300 300 0}, clip]{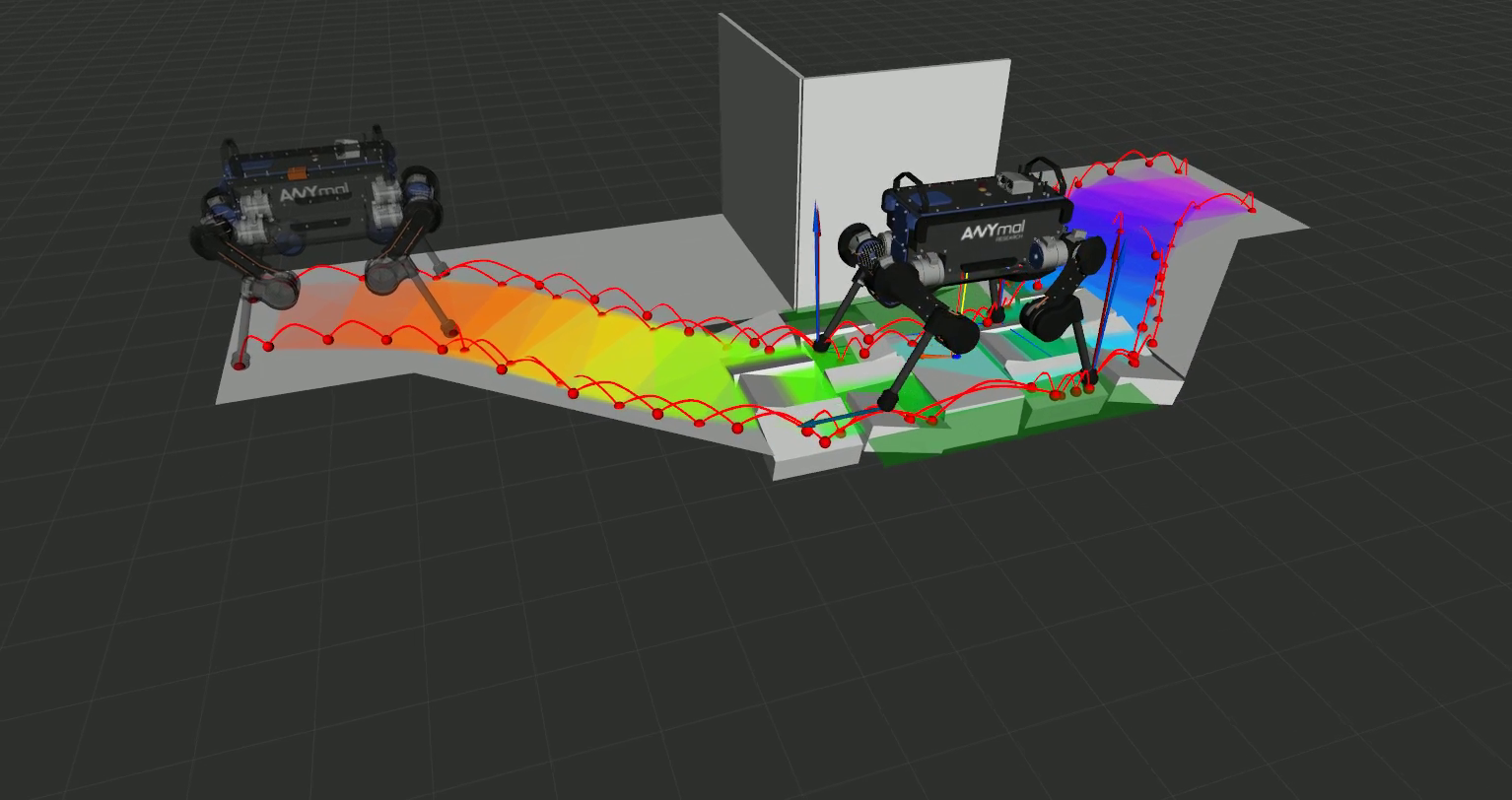}
  \end{subfigure}
  \begin{subfigure}[h!]{0.35\linewidth}
    \includegraphics[width=\linewidth, trim={500 100 500 335}, clip]{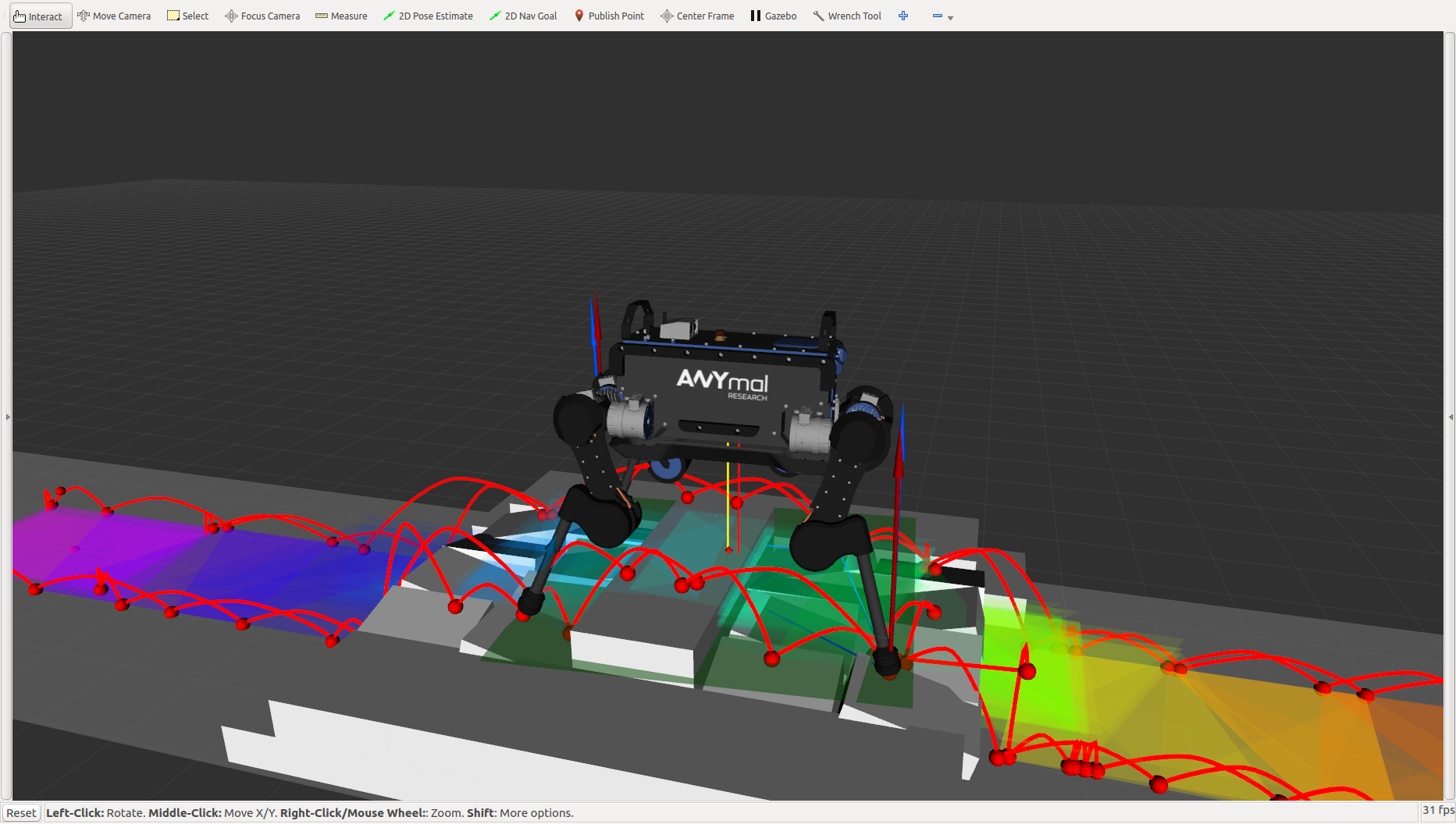}
  \end{subfigure}
  \label{fig:simulation}
  \caption{The ANYmal in dynamical simulation for different environments. Footsteps appear as red spheres while red curves represent the interpolated foot trajectories between consecutive footsteps.}
    \vskip 12pt
\end{figure}

A video showing the trajectories presented in this paper is available via 
\href{https://youtu.be/AliQYvaB78E}{https://youtu.be/AliQYvaB78E}.

\begin{table}[tb]
\centering
\begin{tabular}{|l||c|c|c|c|c|}
\hline
Environments & Affordance & Root & Contact (Number of steps)   & Success \\ \hline \hline
Flat ground (\SI{5}{\metre})  & \SI{1.36}{\milli\second}         & \SI{0.54}{\second}                  & \SI{3.30}{\second} (62.2)                   & 20/20 \\ \hline
Slalom debris    & \SI{8.85}{\milli\second}         & \SI{0.82}{\second}                  & \SI{5.48}{\second} (79.1)                   & 18/20 \\ \hline
Plinth           & \SI{2.27}{\milli\second}         & \SI{1.26}{\second}                   & \SI{4.26}{\second} (59.8)                   & 12/20 \\ \hline
DARPA            & \SI{25.4}{\milli\second}         & \SI{2.18}{\second}                  & \SI{8.94}{\second} (29.9)                  & 7/20 \\ \hline

\end{tabular}
\vskip 3pt
\caption{Mean computation times and success in dynamic simulation, evaluated over 20 runs for each environment.}
\label{tab:computation_time}
\end{table}

\section{Conclusion}

We demonstrated how the HPP planning pipeline can be adapted to automatically generate trajectories for our ANYmal quadruped robot, on a variety of challenging terrains.
We showed that the generated motion plans can be validated on a physically
realistic simulation, and outlined the challenges that can cause
execution to fail. The generation of the full contact plan on average takes
less than \SI{7}{\second} for approximately 50 steps on an environment with many surfaces, on commodity hardware. This makes this algorithm a suitable choice for online replanning in a receding horizon manner.

However, the success rate in dynamic simulation is still too low to allow for unsupervised deployment on the real robot. This problem is primarily a result of the controller. The controller used to generate the whole-body motion and control the center-of-mass motion is too restrictive and
sometimes fails to find a trajectory to link the sequence of static configurations. More specifically, the controller only computes quasi-static trajectories that often reach the limit of the support polygon or that generate high torques. Using a more advanced controller, with a prediction horizon \cite{Carpentier2018,Ponton2016a}, would allow us to compute dynamic motions of the CoM and result to more robust execution of the contact plans.

In future work, we aim to use this planning pipeline on the real ANYmal 
quadruped robot in a set of benchmark examples similar to the environments
presented in this work.





%
%
%
\bibliographystyle{splncs04}
\bibliography{bibliography,drs}

\begin{thebibliography}{10}
\providecommand{\url}[1]{\texttt{#1}}
\providecommand{\urlprefix}{URL }
\providecommand{\doi}[1]{https://doi.org/#1}

\bibitem{Kalakrishnan2010}
Kalakrishnan, M., Buchli, J., Pastor, P., Mistry, M., Schaal, S.: {Learning,
  planning, and control for quadruped locomotion over challenging terrain}. The
  International Journal of Robotics Research  (Nov 2010)

\bibitem{Carpentier2018}
Carpentier, J., Mansard, N.: {Multi-contact Locomotion of Legged Robots}. IEEE
  Transaction on Robotics  (May 2018)

\bibitem{Ponton2016a}
Ponton, B., Herzog, A., Schaal, S., Righetti, L.: {A Convex Model of Momentum
  Dynamics for Multi-Contact Motion Generation}. In: IEEE-RAS International
  Conference on Humanoid Robots (2016)

\bibitem{Herdt2010a}
Herdt, A., Diedam, H., Wieber, P.b., Dimitrov, D., Mombaur, K., Diehl, M.:
  {Online Walking Motion Generation with Automatic Foot Step Placement}.
  Advanced Robotics  (Mar 2010)

\bibitem{Naveau2016}
Naveau, M., Kudruss, M., Stasse, O., Kirches, C., Mombaur, K., Souères, P.: A
  reactive walking pattern generator based on nonlinear model predictive
  control. IEEE Robotics and Automation Letters  (Jan 2016)

\bibitem{2017icra_mastalli}
Mastalli, C., Focchi, M., Havoutis, I., Radulescu, A., Calinon, S., Buchli, J.,
  Caldwell, D.G., Semini, C.: Trajectory and foothold optimization using
  low-dimensional models for rough terrain locomotion. In: IEEE-RAS
  International Conference on Robotics and Automation (2017)

\bibitem{Winkler2018}
Winkler, A., Bellicoso, D., Hutter, M., Buchli, J.: Gait and trajectory
  optimization for legged systems through phase-based end-effector
  parameterization. IEEE Robotics and Automation Letters  (May 2018)

\bibitem{Mastalli2016ICRA}
Mastalli, C., Havoutis, I., Focchi, M., Caldwell, D.G., Semini, C.:
  {Hierarchical Planning of Dynamic Movements without Scheduled Contact
  Sequences}. In: {IEEE-RAS International Conference on Robotics and
  Automation} (2016)

\bibitem{Posa2016}
Posa, M., Kuindersma, S., Tedrake, R.: {Optimization and Stabilization of
  Trajectories for Constrained Dynamical Systems}. IEEE/RAS International
  Conference on Robotics and Automation  (2016)

\bibitem{Mordatch2012}
Mordatch, I., Todorov, E., Popovi{\'{c}}, Z.: {Discovery of complex behaviors
  through contact-invariant optimization}. ACM Transactions on Graphics  (Jul
  2012)

\bibitem{Deits2014}
Deits, R., Tedrake, R.: {Footstep Planning on Uneven Terrain with Mixed-Integer
  Convex Optimization}. In: IEEE-RAS International Conference on Humanoid
  Robots (2014)

\bibitem{Cabezas2018}
Aceituno~Cabezas, B., Mastalli, C., Dai, H., Focchi, M., Radulescu, A.,
  G~Caldwell, D., Cappelletto, J., Grieco, J., Fernandez, G., Semini, C.:
  Simultaneous contact, gait and motion planning for robust multi-legged
  locomotion via mixed-integer convex optimization. IEEE Robotics and
  Automation Letters  (Jan 2018)

\bibitem{Kuffner2001}
Kuffner, J., Nishiwaki, K., Kagami, S., Inaba, M., Inoue, H.: Footstep planning
  among obstacles for biped robots. In: IEEE/RSJ International Conference on
  Intelligent Robots and Systems (2001)

\bibitem{Perrin2013}
Perrin, N., Stasse, O., Lamiraux, F., Yoshida, E.: {Humanoid motion generation
  and swept volumes: theoretical bounds for safe steps}. {Advanced Robotics}
  (Jun 2013)

\bibitem{Winkler2015ICRA}
Winkler, A., Mastalli, C., Havoutis, I., Focchi, M., Caldwell, D., Semini, C.:
  {Planning and execution of dynamic whole-body locomotion for a hydraulic
  quadruped on challenging terrain}. In: IEEE-RAS International Conference on
  Robotics and Automation (2015)

\bibitem{Holden2017}
Holden, D., Komura, T., Saito, J.: Phase-functioned neural networks for
  character control. ACM Transactions on Graphics  (Jul 2017)

\bibitem{deepmind}
DeepMind: Producing flexible behaviours in simulated environments (2017),
  \url{https://deepmind.com/blog/producing-flexible-behaviours-simulated-environments/}

\bibitem{Hwangbo2019}
Hwangbo, J., Lee, J., Dosovitskiy, A., Bellicoso, D., Tsounis, V., Koltun, V.,
  Hutter, M.: Learning agile and dynamic motor skills for legged robots.
  Science Robotics  (Jan 2019)

\bibitem{Tonneau2018}
Tonneau, S., Del~Prete, A., Pettr{\'e}, J., Park, C., Manocha, D., Mansard, N.:
  {An efficient acyclic contact planner for multiped robots}. {IEEE
  Transactions on Robotics}  (Jun 2018)

\bibitem{anymal2016}
Hutter, M., Gehring, C., Jud, D., Lauber, A., Bellicoso, D., Tsounis, V.,
  Hwangbo, J., Bodie, K., Fankhauser, P., Bloesch, M., Diethelm, R., Bachmann,
  S., Melzer, A., Hoepflinger, M.: Anymal - a highly mobile and dynamic
  quadrupedal robot. In: IEEE/RSJ International Conference on Intelligent
  Robots and Systems (2016)

\bibitem{Bouyarmane2009}
Bouyarmane, K., Escande, A., Lamiraux, F., Kheddar, A.: Potential field guide
  for humanoid multicontacts acyclic motion planning. In: IEEE International
  Conference on Robotics and Automation (2009)

\bibitem{Fernbach2017}
Fernbach, P., Tonneau, S., Del~Prete, A., Ta\"ix, M.: {A Kinodynamic
  steering-method for legged multi-contact locomotion}. In: IEEE/RSJ
  International Conference on Intelligent Robots and Systems (2017)

\bibitem{Fernbach2018}
Fernbach, P., Tonneau, S., Ta\"ix, M.: {CROC: Convex Resolution Of Centroidal
  dynamics trajectories to provide a feasibility criterion for the multi
  contact planning problem}. In: IEEE/RSJ International Conference on
  Intelligent Robots and Systems (2018)

\bibitem{hpp}
Mirabel, J., Tonneau, S., Fernbach, P., Sepp{\"a}l{\"a}, A.K., Campana, M.,
  Mansard, N., Lamiraux, F.: {HPP: a new software for constrained motion
  planning}. In: {IEEE/RJS International Conference on Intelligent Robots and
  Systems} (2016)

\bibitem{hpp-link}
LAAS-CNRS: {Humanoid Path Planner}.
  \url{https://humanoid-path-planner.github.io/hpp-doc/download.html?branch=rbprm}
  (2018)

\bibitem{freegait}
Fankhauser, P., Bellicoso, D., Gehring, C., Dube, R., Gawel, A., Hutter, M.:
  Free gait – an architecture for the versatile control of legged robots. In:
  IEEE-RAS International Conference on Humanoid Robots (2016)

\end{thebibliography}

\end{document}